\pdfoutput=1

\documentclass{article}

\usepackage{arxiv}

\usepackage[utf8]{inputenc} % allow utf-8 input
\usepackage[T1]{fontenc}    % use 8-bit T1 fonts
\usepackage{url}            % simple URL typesetting
\usepackage{booktabs}       % professional-quality tables
\usepackage{amsfonts}       % blackboard math symbols
\usepackage{nicefrac}       % compact symbols for 1/2, etc.
\usepackage{microtype}      % microtypography
\usepackage{lipsum}		% Can be removed after putting your text content
\usepackage{graphicx}
\usepackage{natbib}
\usepackage{doi}

\usepackage{framed,multirow}

\usepackage{amssymb}
\usepackage{latexsym}

\usepackage{url}
\usepackage{xcolor}
\usepackage{tabularx}
\usepackage{multirow}
\usepackage{graphicx}

\usepackage{array}    % \extrarowheight macro
\usepackage{booktabs} % for \toprule, \midrule, \bottomrule
\usepackage{pifont}
\usepackage[title]{appendix}

\usepackage{tikz}
\usetikzlibrary{shapes.geometric, arrows}

\definecolor{newcolor}{rgb}{.8,.349,.1}

\title{Domain shifts in dermoscopic skin cancer datasets: Evaluation of essential limitations for clinical translation}

\date{} 					% Or removing it
  \author{Katharina Fogelberg\thanks{\texttt{Both authors contributed equally}}\\
	Digital Biomarkers for Oncology \\
	German Cancer Research Center (DKFZ)\\
	Heidelberg, Germany \\
	\texttt{katharina.fogelberg@dkfz-heidelberg.de} \\
	\And
	Sireesha Chamarthi\footnotemark[1]\\
	Data Analysis and Intelligence\\
	German Aerospace Center (DLR)\\
	Jena, Germany \\
	\texttt{Sireesha.Chamarthi@dlr.de} \\
	\AND
	Roman C. Maron \\
        Digital Biomarkers for Oncology \\
	German Cancer Research Center (DKFZ) \\
	\And
	Julia Niebling \\
        Data Analysis and Intelligence\\
	German Aerospace Center (DLR) \\
	\And
	Titus J. Brinker \\
        Digital Biomarkers for Oncology \\
	German Cancer Research Center (DKFZ) \\
}

\renewcommand{\headeright}{Fogelberg et al, 2023}
\renewcommand{\undertitle}{Preprint}
\renewcommand{\shorttitle}{Domain Shifts in Dermoscopic Datasets}

\hypersetup{
pdftitle={Domain shifts in dermoscopic skin cancer datasets: Evaluation of essential limitations for clinical translation},
pdfsubject={cs.CV},
pdfauthor={Katharina Fogelberg, Sireesha Chamarthi},
pdfkeywords={domain shift, skin lesion classification},
}

\begin{document}
\maketitle

\begin{abstract}
The limited ability of Convolutional Neural Networks to generalize to images from previously unseen domains is a major limitation, in particular, for safety-critical clinical tasks such as dermoscopic skin cancer classification. In order to translate CNN-based applications into the clinic, it is essential that they are able to adapt to domain shifts. Such new conditions can arise through the use of different image acquisition systems or varying lighting conditions. In dermoscopy, shifts can also occur as a change in patient age or occurence of rare lesion localizations (e.g. palms). These are not prominently represented in most training datasets and can therefore lead to a decrease in performance. In order to verify the generalizability of classification models in real world clinical settings it is crucial to have access to data which mimics such domain shifts. To our knowledge no dermoscopic image dataset exists where such domain shifts are properly described and quantified. We therefore grouped publicly available images from ISIC archive based on their metadata (e.g. acquisition location, lesion localization, patient age) to generate meaningful domains. To verify that these domains are in fact distinct, we used multiple quantification measures to estimate the presence and intensity of domain shifts. Additionally, we analyzed the performance on these domains with and without an unsupervised domain adaptation technique. We observed that in most of our grouped domains, domain shifts in fact exist. Based on our results, we believe these datasets to be helpful for testing the generalization capabilities of dermoscopic skin cancer classifiers. 
\end{abstract}

% keywords can be removed
\keywords{domain shift \and skin lesion classification \and dermoscopic image \and unsupervised domain adaptation \and generalization \and clinical translation}

\newpage

\section{Introduction}
\label{sec1}
% 1. problem statement: domain shifted datasets needed to test generalization of lesion classification models

Convolutional Neural Networks (CNNs) in dermoscopic skin lesion classification tasks have been shown to be on par or even outperform dermatologists in experimental settings \citep{esteva2017dermatologist, brinker2019deep, maron2019systematic, haenssle2021skin}. Nevertheless, due to generalization issues, these models are not yet ready to be translated into the clinic \citep{oloruntoba2022assessing}. In conventional Computer Vision (CV) tasks, domain shifts (new conditions) are rarely present, which means that the source (training) and target (testing) datasets are drawn from the same data distribution. However, in real world scenarios that is usually not the case. In general, CNNs have limited capabilities to adapt to domain shifts, thus resulting in a decrease in performance \citep{torralba2011unbiased, stacke2020measuring, ouyang2022causality}. Insufficient generalization in safety-critical medical applications can potentially jeopardize patient safety. Although various methods have been proposed to tackle the generalization challenge, there is currently an inadequate amount of testing data available to verify that these methods are effective for medical image data, such as domain shifted dermoscopic images. Recent research in the field of pathology highlights the significance of appropriately created datasets for assessing the performance of CNN models \citep{AI_pathology2020, Homeyer2022}. As the use of CNNs in medical settings becomes increasingly common, it is important to implement reliable and trustworthy models and enhance confidence in their results \citep{Mller2022}. The creation of accurate and practical real-life datasets is a crucial step towards achieving this objective.

\begin{figure*}[htb]
\begin{minipage}[b]{1.0\linewidth}
  \centering
  \centerline{\includegraphics[width=\textwidth]{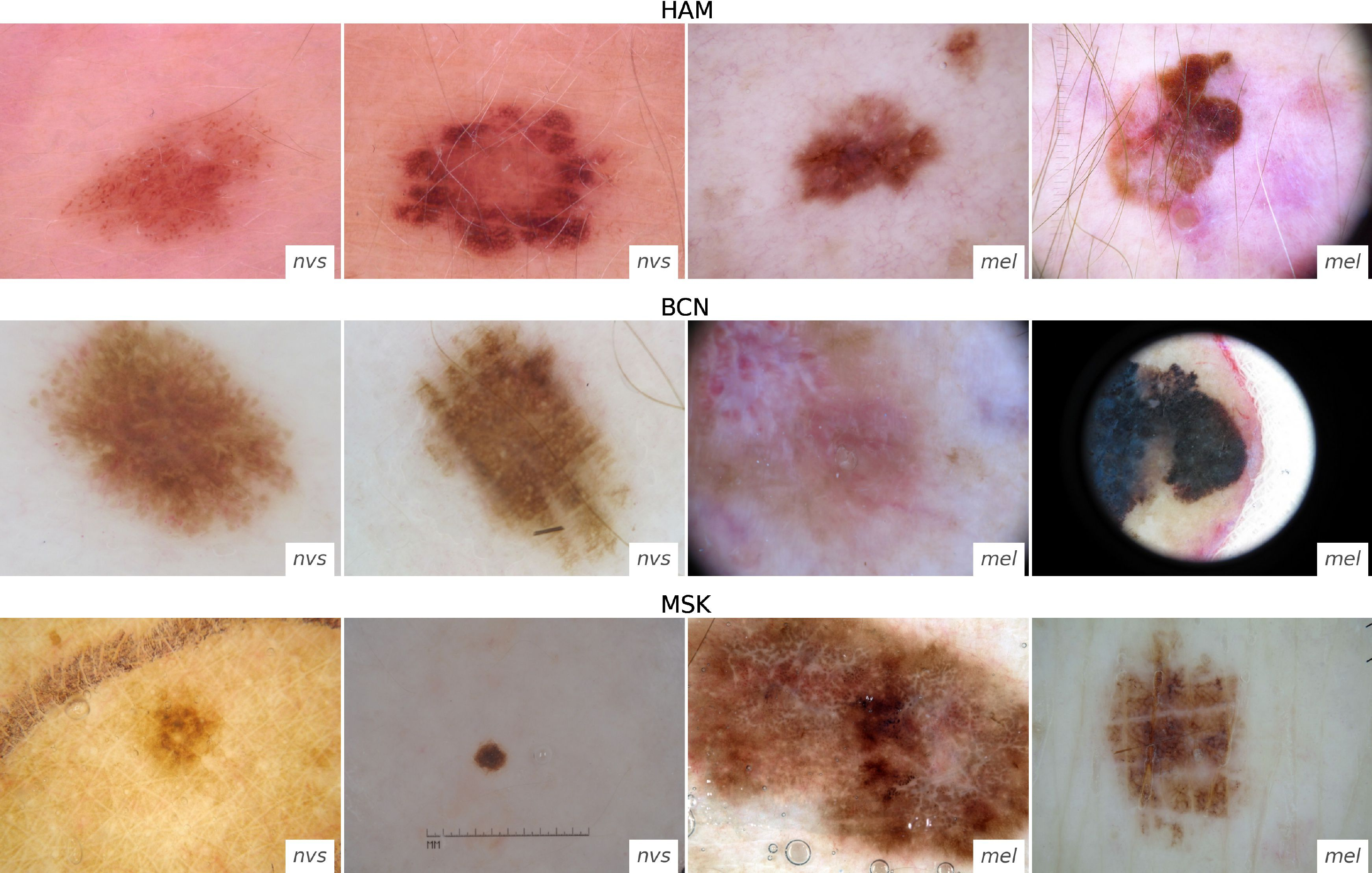}}
  \caption{Randomly selected nevus (nvs) and melanoma (mel) images from the ISIC datasets HAM, BCN and MSK.}
\label{fig:derma}
\end{minipage}
\end{figure*}

% 2. explain domain shifts in dermoscopic images - which are integrated in such public datasets
Typical benchmark data for domain adaptation tasks exists, e.g. digit datasets (MNIST, MNISTM, rotated MNIST, USPS, SVHN) or office objects (OFFICE dataset consists of three domains: Amazon, Webcam, DSLR) \citep{saenko2010adapting}. Different adaptation techniques work well on these datasets \citep{ganin2015unsupervised, ganin2016domain, wang2018deep, wilson2020survey, zhang2021survey}. However, good adaptation in digit images does not guarantee the same in dermoscopic images, where the shifts are difficult to recognize visually (\autoref{fig:derma}). For skin cancer classification tasks, there exist various sources which provide publicly available dermoscopic images along with their metadata, such as PH2 \citep{mendoncca2013ph}, SKINL \citep{de2019light} and DERM7PT \citep{kawahara2018seven}. The largest and most popular collection of databases comes from the International Skin Imaging Collaboration (ISIC) \citep{rotemberg2021patient} which contains skin lesion images from different clinics. This variety of clinics and the way these datasets are published makes it difficult to create a standardized environment for researchers to make their results reproducible.

To our knowledge, none of these dermoscopic datasets are assembled to multiple domains which are designed to test the generalization capabilities of skin lesion classification models. For the various publicly available dermoscopic images it can only be speculated how large domain shifts between different datasets truly are and what presumably could cause these shifts: technical or biological change? Also, there has been limited effort in understanding and quantifying the domain shift between and within datasets \citep{venkateswara2017deep}. Domain shifts are either visually identifiable or are simply assumed to exist.
% 3. also there has been only limited effort to quantify such domain shifts - need for universal metric?

% 4. what can be solutions to handle such domain shifted datasets to still have good performances? domain adaptation techniques, for real world scenarios/translation and the lack of medical data availability especially unsupervised kind. Some related work?
For a successful translation of such diagnostic Artificial Intelligence (AI) systems into the clinic, approaches to properly handle domain shifts are needed to generalize well on unseen data. For this purpose, multiple methods, such as image normalization, data augmentation and domain adaptation can be used. In this work we want to focus on domain adaptation techniques to improve the performance of our classifier on the grouped domain shifted test sets. As a major limitation in the medical field is the lack of labelled data, especially unsupervised domain adaptation techniques are the ones of interest. There have been advancements in the field of medical image classification using unsupervised domain adaptation (UDA) methods \citep{ren2018adversarial, ren2019unsupervised, gu2019progressive, guan2021domain}.\\

%\newpage
% our contribution: providing new domain shifted datasets to test generalization abilities of your skin lesion classification models. Try to come up with a relevant quantification measure - which one could be suitable? And showing the impact of the intensity of domain shift on one uda method
Our main contributions are summarized to:
\begin{itemize}
    \item Characterizing possible factors which lead to potential domain shifts in real world clinical dermoscopy.
    \item Quantifying potential domain shifts in between different and within the same dataset based on feature-, image- and visualization-based methods.
    \item Grouping of public dermoscopic images into domain shifted datasets. Our aim is to provide these grouped datasets for other researchers working with dermoscopic image data, so they can test the generalization capabilties of their models. For this we release a repository\footnote{\texttt{https://gitlab.com/dlr-dw/isic\char`_download}} to automatically download the domain shifted datasets.
    \item Investigating the influence of domain shifts on one UDA method. We believe that it can give a better insight into the relationship between domain shifts and adaptation techniques. 
\end{itemize}

%paper content
This work is structured into the following parts: The related work discussed in \autoref{sec:Relevant} deals with existing research on domain shifts and highlights our motivation to identify domain shifts in dermoscopic datasets. In \autoref{sec:sampling} we discuss possible domains which can arise in skin lesion classification and our rationale behind splitting the data into different categories. We further verify if our grouped datasets are in fact domain shifted by using multiple quantification metrics in \autoref{sec:domainshift_quantification}. Finally, on a melanoma vs. nevus classification task we compare the performances  of a ResNet50 with and without an unsupervised domain adaptation method on all grouped domain shifted datasets. The methodology and results are presented in \autoref{sec:influence}. We conclude the paper with our findings and discuss possible future research directions in \autoref{sec:discussion}.

\section{Related work}
\label{sec:Relevant}

\label{ssec:isic}

%2. discussion about the references so far for similarity and dissimilarity metrics
When focusing on domain shifts, one dataset does not necessarily have to represent only one domain \citep{plank2011effective}. In some cases, it is also possible that multiple sub-domains exist within the same dataset or that two different datasets are considered as only one domain. A domain shift in dermoscopic datasets can be caused by different factors which can be categorized into groups. \cite{finlayson2021clinician}, grouped clinical datasets into shifts by technology, population and behaviour. In this paper we want to take a slightly different direction focusing on dermoscopic data shifts specifically, where we mainly group the data into technical and biological shifts. Different image acquisition system settings like brightness and contrast or lighting conditions are designated as technical domain shift. On the other hand, biological domain shifts can be caused by variations in the lesion localization and different age groups of patients. However, for the existing dermoscopic datasets these shifts are not described and quantified, but rather just assumed to exist or completely disregarded. Therefore a quantification of the intensity of these differences is needed to properly test the ability of classification models to generalize across domains. 

Unfortunately there is no universal quantification measure. Divergence measures are widely used to quantify domain shifts across CV- \citep{omer2006image,zhou2020learning} and NLP- \citep{kashyap2020domain} tasks, as well as in the medical field \citep{stacke2020measuring, ionescu2022similarity}. 

%Most distance measures, which are used to quantify domain shifts are utilized within the Natural Language Processing (NLP) context \citep{ramesh2021domain}, but there are also a few in CV tasks \citep{omer2006image} and even in the medical field \citep{stacke2020measuring, ionescu2022similarity}. 

%In the survey of \cite{ramesh2021domain} the authors categorize divergence measures into three main classes. Geometric measures are measured in the metric space by calculating the distance between feature vectors. Information-theoretic measures focus on the distance between probability distributions. Higher-Order measures also focus on the distance between distributions, but in the feature space. This categorization is from a NLP focused paper. 

% representation shift 
\cite{stacke2020measuring} estimated the mean divergence between the source and target datasets in the model specific latent representation space. It is calculated over all convolutional filters while discarding irrelevant features for the task to limit the model to learn only features that might be very specific to the domain.
The authors used three types of divergence metrics: Wasserstein distance, Kullback-Leibler (KL) divergence \citep{van2014renyi} and Kolmogorov Smirnov Statistic  to measure the so-called “representation shift”. Also, they state that between their representation shift and accuracy there is higher correlation using Pearson correlation coefficient than with other approaches. The larger the representation shift, the higher the risk of performance degradation. Due to the quantification on feature-level, the limitation of this metric is that it is tightly connected to the specific model.

On feature-level, other works show that also Cosine similarity, for instance, can be used to measure the shift by calculating the distance between two image vectors in the feature space \citep{wang2020cross}. Besides, a KL divergence-based approach was used to compare the similarity between medical images \citep{pheng2016kullback}. 

% 4. Reason why we also tried image level simialrity
There is also research based on image-specific calculations of similarity between pixel values \citep{omer2006image}. \cite{palladino2020unsupervised} used only the Jensen-Shannon (JS) divergence to calculate the average difference between two probability distributions. They calculated it pairwise on image-level, between all possible pairs of images from two domains, intra- and inter-domain. Also they show an anti-correlation between JS divergence and their performance measure: lower inter-domain divergence results in better generalization from source to target. In comparison to \cite{stacke2020measuring}, the limitation is that the quantification only happens in image space by using only one divergence measure.
%5. what information does domain classifier gives

\cite{rabanser2019failing} combined statistical hypothesis testing with dimensionality reduction to detect domain shifts. The objective of their work is to differentiate between small and large domain shifts present in the data. For that, they evaluated the most similar and dissimilar images returned by a domain discriminator.

Overall, works applying domain shift quantification use various approaches while focusing on distinct details. Based on the absence of a universal measure, it could be incomplete to use only one quantification approach. However, what all works have in common is that they use some kind of divergence measure. Furthermore, they all show a correlation or anti-correlation between the divergence metric and the performance (drop) of the classifier.

Apart from this, another popular method for visualizing feature distributions of different domains in datasets is t-distributed Stochastic Neighbor Embedding (t-SNE) \citep{van2008visualizing, ganin2016domain}. t-SNE projections are widely used in domain adaptation applications for visualizing high dimensional data in 2D and for estimating the domain separations. In domain adaptation applications, results have proven that there is a correlation between performance of the adaptation methods and the overlap between domains \citep{ganin2016domain}. In our analysis, we used t-SNE projections to estimate the domain shifts in the datasets.

A further way to estimate the presence of domain shifts in datasets is by measuring the performance drop in a classification task between the source and the target sets \citep{elsahar2019annotate, stacke2020measuring}. This is achieved by a domain classifier which should be able to easily distinguish datasets from different domains, where a large domain shift exists in between them \citep{stacke2020measuring}. However, due to the presence of duplicates in the datasets (except for HAM), a domain discriminator cannot be used as a reliable metric to quantify domain shifts in ISIC datasets \citep{cassidy2022analysis}. Consequently, we did not use a domain discriminator as a primary quantification metric, but we correlated it with our quantification results in \autoref{sec:discussion}.

Divergence measures like KL and JS divergence, as well as Cosine similarity are used in different CV-tasks by focusing on either feature- or image-level measurements. Areas which utilize these metrics are similarity detection \citep{chen2020exploring}, image segmentation \citep{katatbeh2015optimal}, object identification and face verification \citep{nguyen2010cosine}, contrastive learning \citep{rezaei2021deep} and quality assurance \citep{bruni2013jensen}.  In the medical field specifically, such measures are mostly used in areas like image segmentation \citep{sandhu2008new,taha2015metrics} and image registration \citep{martin2007new}.

\section{Grouping potential domain shifted datasets from ISIC archive}
\label{sec:sampling}

To test the generalization capabilities of a skin lesion classifier or to prove how effective a UDA technique is in adapting images to new domains, it is important to have appropriate datasets from different domains. 

\subsection{ISIC archive and domain shifts}

ISIC archive is an open-source collection of databases from clinics across the globe, which contains over 69000 dermoscopic images of benign and malignant skin lesions with corresponding patient metadata  \citep{rotemberg2021patient}. For the data grouping process we selected three large datasets from ISIC archive: HAM \citep{tschandl2018ham10000}, BCN \citep{combalia2019bcn20000} and MSK \citep{cassidy2022analysis}, which are all also available on separate websites. HAM (Human Against Machine) comprises 10000 skin lesion images including a lesion ID to identify duplicate images. BCN is a collection of lesion images from facilities of the Hospital Clinic in Barcelona collected from 2010 to 2016. MSK consists of lesion images collected at the Memorial Sloan-Kettering Cancer Center in the US. \autoref{fig:derma} shows melanoma and nevus images from each of the datasets. It can be noted that it is challenging to visually distinguish between the different domains present in the images. Apart from that, \autoref{fig:derma} also demonstrates the presence of artefacts, as rulers, markers and black borders included in the images, which can highly affect performance results of a classifier. 

\begin{figure}[htb]
\begin{minipage}[b]{0.5\textwidth}
  \centering
  \centerline{\includegraphics[width=\textwidth]{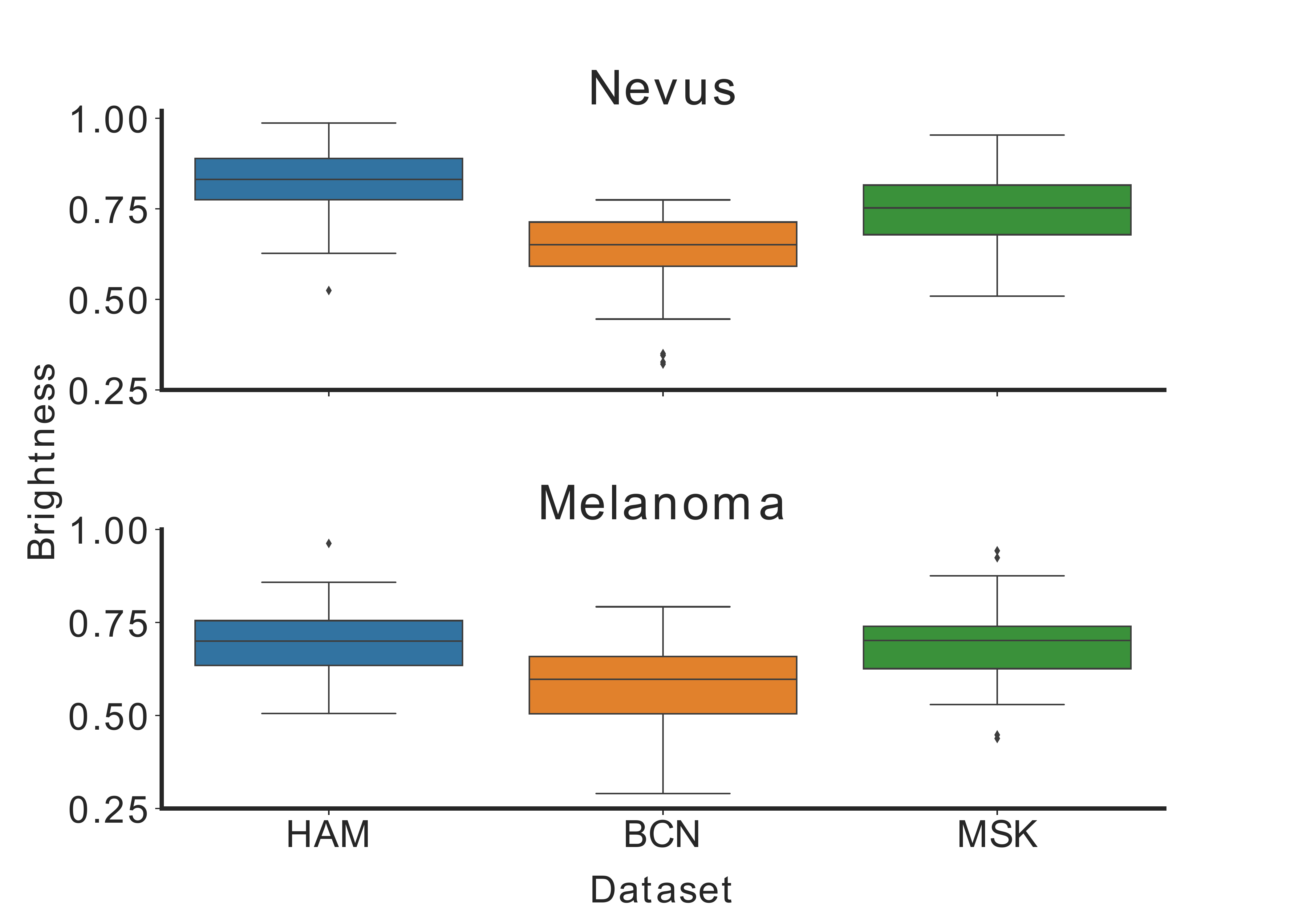}}
  \medskip
\end{minipage}
\begin{minipage}[b]{0.5\textwidth}
  \centering
  \centerline{\includegraphics[width=\textwidth]{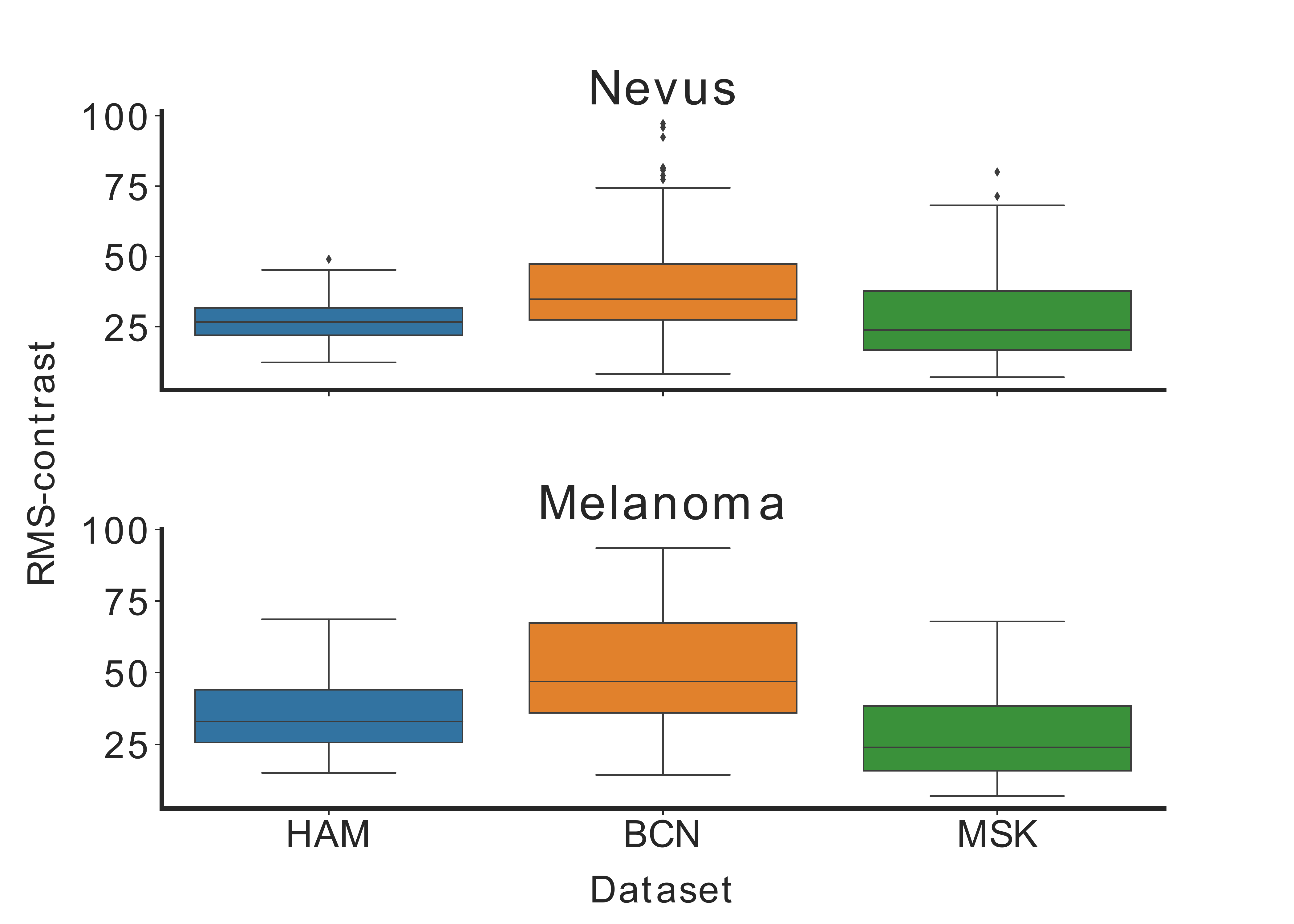}}
\end{minipage}
\caption{Box plots of image properties (brightness, rms-contrast) in HAM, BCN and MSK datasets, which show differences on image-level.}
\label{fig:techn-shift1}
\end{figure}

The images present within each of the datasets are obtained from a wide range of patients and can include multiple lesions of the same patient taken at different points in time. The differences in the images are chiefly caused due to the variation in the a) skin lesions from different patients and b) mechanism/device that captures the image. In this dermoscopic use case we categorise these two as potential technical and biological shifts. 

\begin{figure}[htb]
\begin{minipage}[b]{0.5\textwidth}
  \centering
  \centerline{\includegraphics[width=\textwidth]{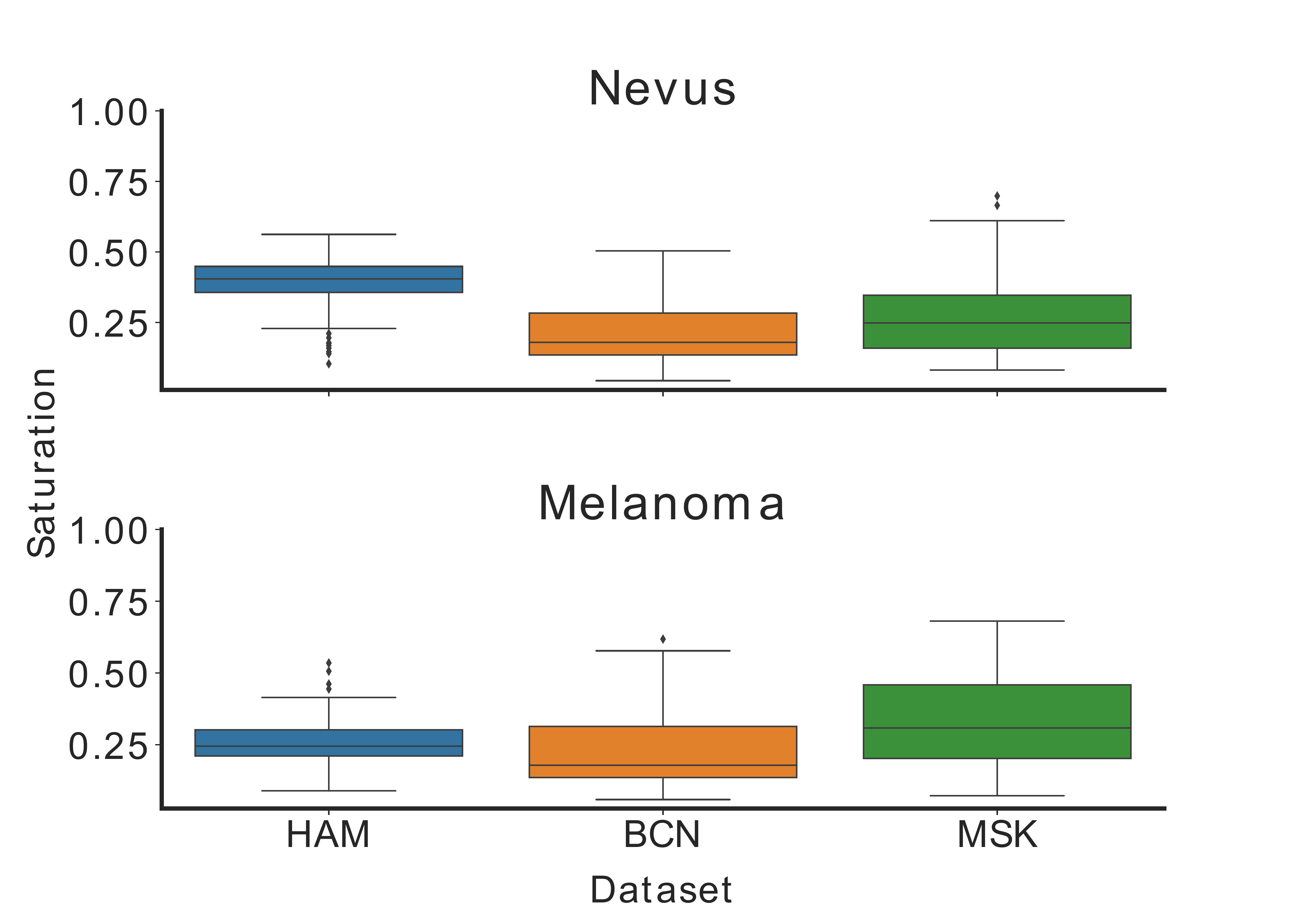}}
  \medskip
\end{minipage}
\begin{minipage}[b]{0.5\textwidth}
  \centering
  \centerline{\includegraphics[width=\textwidth]{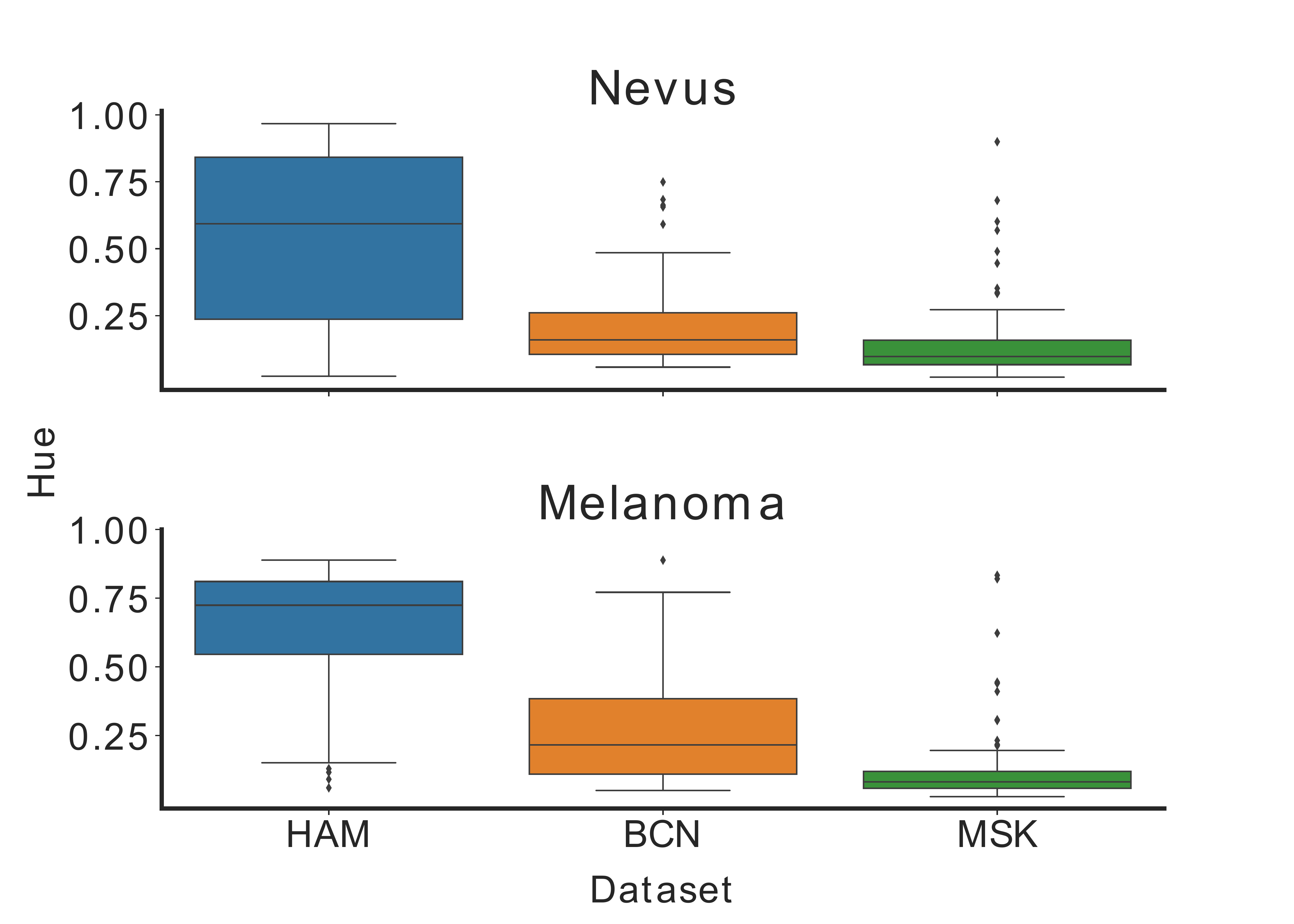}}
\end{minipage}
\caption{Box plots of image properties (saturation, hue) in HAM, BCN and MSK datasets, which show differences on image-level.}

\label{fig:techn-shift2}
\end{figure}
\begin{figure}[h!]
\begin{minipage}[b]{0.5\textwidth}
  \centering
  \centerline{\includegraphics[width=\textwidth]{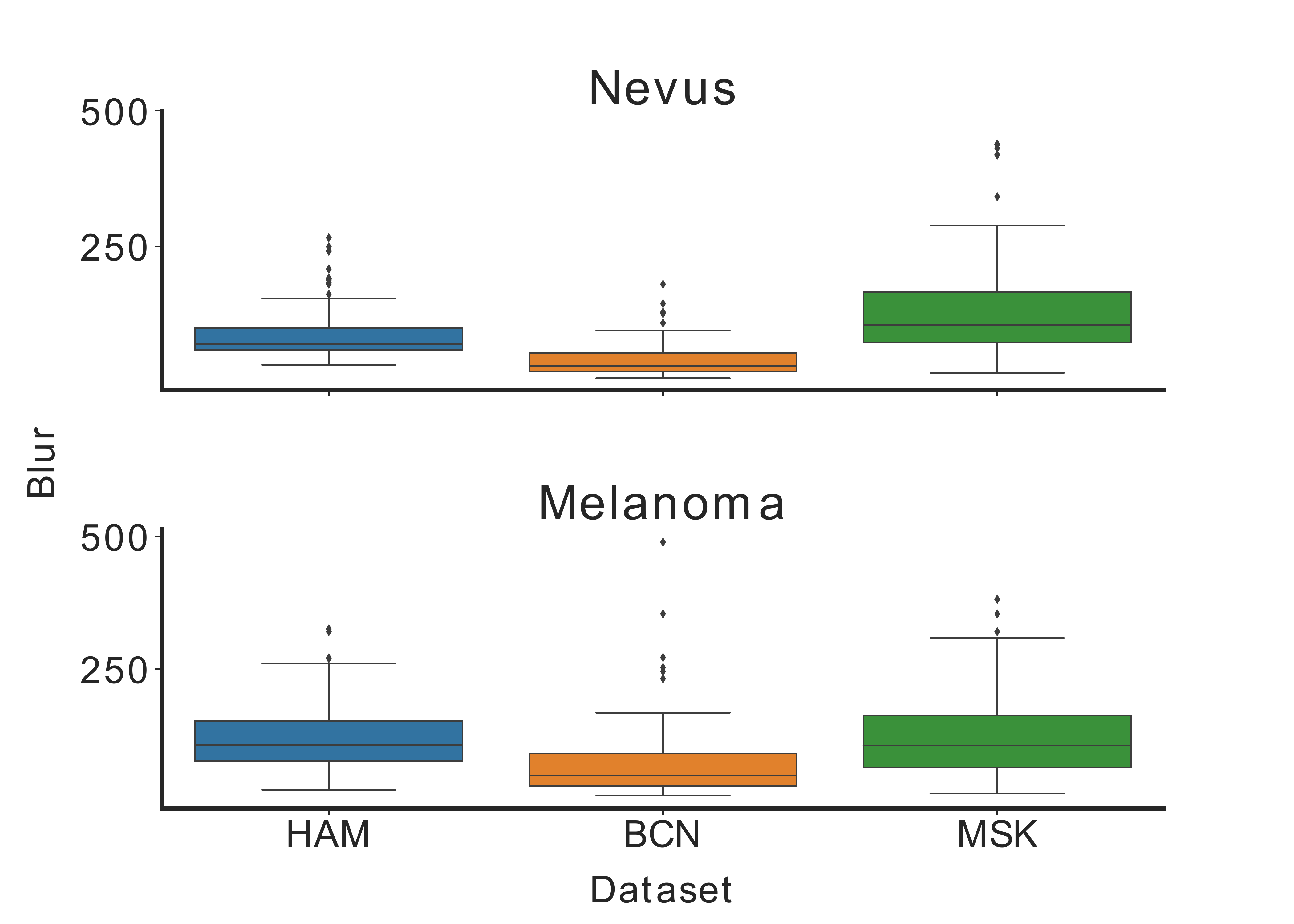}}
\end{minipage}
\caption{Box plots of image properties (blur - Laplacian of HSV color space) in HAM, BCN and MSK datasets, which show differences on image-level. Outliers were excluded from the plot.}
\label{fig:techn-shift3}
\end{figure}

\begin{itemize}
    \item Technical shifts: as HAM, BCN and MSK were acquired in different clinics and  countries, different technical settings can be expected. Actual information about the acquisition systems capturing the images is not available in the metadata of these datasets. Dermoscope settings and lighting conditions in the clinic, for instance, can cause the dermoscopic images to look differently. To confirm our assumption about differences in the image properties of the three datasets, we analyzed and compared brightness and RMS (Root Mean Squared)-contrast (\autoref{fig:techn-shift1}), saturation and hue (\autoref{fig:techn-shift2}) and blur (\autoref{fig:techn-shift3}) of their images. As melanoma images may look darker overall, we analyzed the technical properties separately for the classes melanoma and nevus. \\
    We can observe, that while using only a subset of the entire dataset (n=450 per class and dataset), there is already a clear difference in between the image properties of the three datasets. We undersampled by the dataset class of the smallest size. Unexpectedly, melanoma and nevus do not show large variations in image properties. The technical differences in the images may have an influence on the performance of a model trying to classify one of the three datasets and training on another. Consequently, we focused on the origin of the datasets as technical shifts.
    the three represented datasets represent a technical shift.
    \item Biological shifts: arise due to multiple patient- and lesion-related factors. As the dermoscopic images are collected from various patients, there will be a range of different skin-types, -colors and -lesions recognizable within the images. Another relevant factor is age, as age groups older than 30 typically have a higher risk of having melanoma \citep{paulson2020age}. As melanoma in men occurs as frequently as in women \citep{paulson2020age}, it is most likely not a relevant factor for domain shifts. Recent works confirmed that gender does not correlate with the diagnosis \citep{hohn2021integrating, sies2022does}. Additionally, a complete male and female split would never happen in a real world setting, although the preferred melanoma localization differs by gender. Melanoma in men occurs mostly on the torso, while in women melanoma are mainly diagnosed in the lower extremities \citep{stanienda2017primary}. The reason is probably the extent of sun-exposure in these body localizations, as these have a higher risk for melanoma \citep{stanienda2017primary}. Localization may represent a domain shift factor, because lesions localized at the torso and lesions localized at the palms and soles will look different, as well as their surrounding skin type. Taking into account the factors discussed above, we focused on age and localization as biological shifts in our grouping procedure.
\end{itemize}

\subsection{Grouping procedure}
\label{ssec:sampling_procedure}
%flowchart style
\tikzstyle{Source} = [circle, rounded corners, minimum width=0.7cm, minimum height=0.7cm,text centered, draw=black]
\tikzstyle{Target} = [circle, rounded corners, minimum width=0.7cm, minimum height=0.7cm,text centered, draw=black]
\tikzstyle{Target2} = [circle, rounded corners, minimum width=0.7cm, minimum height=0.7cm,text centered, draw=black]
\tikzstyle{process} = [rectangle, minimum width=2cm, minimum height=1cm, text centered, draw=black, fill=white!30]
\tikzstyle{String} = [minimum width=2cm, minimum height=1cm, text centered, fill=white!30]
\tikzstyle{process2} = [rectangle, minimum width=1cm, minimum height=1cm, text centered, draw=black, fill=white!30]
\tikzstyle{decision} = [diamond, minimum width=1.3cm, minimum height=0.7cm, text centered, draw=black, fill=white!30]
\tikzstyle{edge} = [thick]
%flowchart
\begin{figure}[h!]
\centering
\resizebox{7cm}{12cm}{
\begin{tikzpicture}[node distance=1.7cm]
\node (start)[process] {ISIC Dataset};
\node (pro2) [process, below of=start, xshift=-3cm] {HAM};
\node (pro3) [process, below of=start] {BCN};
\node (pro4) [process, below of=start, xshift=3cm] {MSK};
\node (circ1) [Target, below of=start, yshift=-50] {Target};
\node (circ2) [Target, below of=start, xshift=3cm, yshift=-50] {Target};
\node (circ3) [Source, below of=start, xshift=-3cm, yshift=-50] {Source};
\node (text1) [String, below of=start, xshift=-3cm, yshift=75] {a) Technical shift};
\node (text2) [String, below of=start, xshift=-3cm, yshift=-95] {b) Biological shift};
\draw [edge] (start) -- (pro3);
\draw [edge] (pro2) -- (circ3);
\draw [edge] (pro3) -- (circ1);
\draw [edge] (pro4) -- (circ2);
\draw [edge] (pro4) |- (start);
\draw [edge] (pro2) |- (start);
\node (Target2) [Target2, below of=circ1, yshift=-20,text width=1cm, xshift=-25] {Source/\\Target};
\node (dec1) [decision, below of=Target2, yshift=-20] {Age $\leq30$};
\draw [edge] (Target2) -- (dec1);
\node (pro1) [process2, below of=Target2, xshift=-2.1cm, text width=1.8cm, yshift=-80] {Age $\leq30$ \\(All localizations)};
\draw [edge] (dec1) -| node[anchor=south, xshift=13] {yes} (pro1);
\node (dec2) [decision, below of=Target2, xshift=60, yshift=-80, text width=0.6cm] {Loc. = Body};
\draw [edge] (dec2) |- node[anchor=south, xshift=-13] {no} (dec1);
\node (pro5) [process2, below of=dec2, xshift=2cm, text width=1.8cm, yshift=-20] {Loc. = Head/Neck};
\node (pro6) [process2, below of=pro5, text width=1.8cm, yshift=16] {Loc. = Palms/Soles};
\node (pro7) [process2, below of=dec2, text width=1.8cm, xshift=-60, yshift=-20] {Loc. = Body (default)};
\draw [edge] (pro5) |- node[anchor=south, xshift=-13] {no} (dec2);
\draw [edge] (pro7) |- node[anchor=south, xshift=13] {yes} (dec2);
\end{tikzpicture}
}
\caption{Overview of the grouping process to generate domain shifted datasets for dermoscopic images. The three datasets (HAM, BCN, MSK) follow the same splitting procedure, where they are grouped by patient age and different lesion localizations. The leaf nodes represent the resulting potential domain shifted datasets.}
\label{fig:sampling_procedure}
\end{figure}
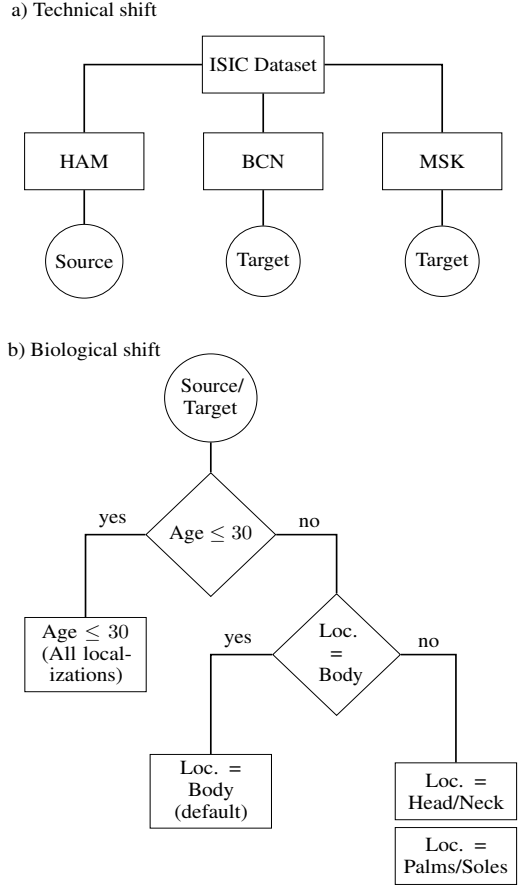

\begin{table*}[h!]
\centering
\begin{tabular}{llllll@{}}
%\multicolumn{2}{c}{}{}\hspace{0.8cm}Abbreviation & Origin & Biological factors & Biological shift & Technical shift  \\
\toprule[0.9pt]
Abbreviation & Origin & Biological factors & Class distribution & Biological shift & Technical shift \\
\toprule[0.9pt]
%\multirow{5}{*}{\rotatebox[origin=c]{90}{MSK\hspace{1.1cm}BCN\hspace{1.1cm}HAM\hspace{0.5cm}}}
H & HAM & Age \textgreater 30, Loc. = Body (default) & $465 : 4234\hspace{0.1cm}(4699)$ &  & \\ [0.2ex]
HA & HAM & Age $\leq30$, Loc. = Body & $25 : 532\hspace{0.1cm}(557)$ & \checkmark & \\ [0.2ex]
HLH & HAM & Age \textgreater 30, Loc. = Head/Neck & $99 : 121\hspace{0.1cm}(220)$ & \checkmark & \\[0.2ex]
HLP & HAM & Age \textgreater 30, Loc. = Palms/Soles & $15 : 203\hspace{0.1cm}(218)$ & \checkmark & \\ [0.2ex]
\midrule
B & BCN & Age \textgreater 30, Loc. = Body (default) & $1918 : 2721\hspace{0.1cm}(4639)$ &   & \checkmark \\ [0.2ex]
BA & BCN & Age $\leq30$, Loc. = Body & $71 : 808\hspace{0.1cm}(879)$ & \checkmark & \checkmark \\ [0.2ex]
BLH & BCN & Age \textgreater 30, Loc. = Head/Neck & $612 : 320\hspace{0.1cm}(932)$ & \checkmark & \checkmark \\[0.2ex]
BLP & BCN & Age \textgreater 30, Loc. = Palms/Soles & $192 : 105\hspace{0.1cm}(297)$ & \checkmark & \checkmark \\ [0.2ex]
\midrule
M & MSK & Age \textgreater 30, Loc. = Body (default) & $565 : 1282\hspace{0.1cm}(1847)$ &  & \checkmark \\ [0.2ex]
MA & MSK & Age $\leq30$, Loc. = Body & $37 : 427\hspace{0.1cm}(464)$ & \checkmark & \checkmark \\ [0.2ex]
MLH & MSK & Age \textgreater 30, Loc. = Head/Neck & $175 : 117\hspace{0.1cm}(292)$ & \checkmark & \checkmark \\[0.2ex]
\bottomrule[0.9pt]
\end{tabular}
\caption{Representation of grouped datasets resulting from HAM, BCN and MSK with its abbreviations. Additionally, dataset sizes and class distributions (melanoma : nevus (total)) are included. H is further split 80:20\hspace{0.1cm}(372 : 3387 (3759), 93 : 847 (940)) to generate a train and holdout set. Potential biological or technical domain shifts are emphasized with checkmarks.}
\label{table:samplingtable1}
\end{table*}

% \begin{table*}[h!]
% \centering
% \label{table:samplingtable2}
% \begin{tabular}{|c|cccc@{}}
% \cmidrule[0.9pt](l){2-4}
% \multicolumn{2}{c}{\hspace{5.1cm}HAM}& BCN & MSK & \\
% \cmidrule(l){2-4}
% \multirow{5}{*}{{}}
% Age \textgreater 30, Loc. = Body (default) & $465 : 4234\hspace{0.1cm}(4699)$ & $1918 : 2721\hspace{0.1cm}(4639)$ & $565 : 1282\hspace{0.1cm}(1847)$ \\ [0.2ex]
% Age $\leq30$, Loc. = Body & $25 : 532\hspace{0.1cm}(557)$ & $71 : 808\hspace{0.1cm}(879)$ & $37 : 427\hspace{0.1cm}(464)$ \\ [0.2ex]
% Age \textgreater 30, Loc. = Head/Neck & $99 : 121\hspace{0.1cm}(220)$ & $612 : 320\hspace{0.1cm}(932)$ & $175 : 117\hspace{0.1cm}(292)$ \\[0.2ex]
% Age \textgreater 30, Loc. = Palms/Soles & $15 : 203\hspace{0.1cm}(218)$ & $192 : 105\hspace{0.1cm}(297)$ & $-$ \\ [0.2ex]
% \cmidrule[0.9pt](l){2-4}
% \end{tabular}
% \caption{Grouped domain adaptation dataset sizes and class distributions (melanoma : nevi (total)). HAM default is split 80:20\hspace{0.1cm}(372 : 3387 (3759), 93 : 847 (940)) to generate train and holdout sets. Columns represent technical shift. Rows represent biological shift.}
% \end{table*}

We grouped the data present in ISIC archive into technical and biological domains as shown in \autoref{fig:sampling_procedure}. This procedure needs to take into account the avoidance of data leakage and the reproducibility for further datasets. The three datasets HAM, BCN and MSK represent the technical shifts as the images are obtained from different clinics and acquisition systems. HAM was used as training and holdout-validation set (source), although it contains fewer images than BCN. As HAM is the only dataset that contains lesion IDs, while all three datasets contain potential duplicates, it is considered an optimal training set to avoid the falsification of performance results due to data leakage. The typical patient with skin cancer is predominantly older than 30 and with a lesion localized in the main body area, including torso, upper- and lower extremities \citep{paulson2020age}. That is why images from patients with these properties are considered as default for our grouping procedure.

\autoref{table:samplingtable1} provides an overview of the resulting grouped datasets from HAM, BCN and MSK. While the rows are clustered into groups from the same dataset, the columns give further information. We emphasized potential domain shifts, which are either biological, technical or both, with checkmarkers. If each of the generated domain shifted datasets can in fact be considered as truly domain shifted has to be verified. Also abbreviations for the resulting domain shifted datasets, which will further be referred in the following sections, can be found in \autoref{table:samplingtable1}. The table also represents dataset sizes and class distributions for each resulting grouped dataset. Total sizes of each dataset are not comparable to the original downloaded sets, as we used only images which are classified as either melanoma or nevi, while excluding other lesion types. Grouped datasets which resulted in a small dataset size ($\leq200$ images in total) were removed from our experiments (e.g. Age $>30$, Loc. = Oral/Genital with only 19 melanomas and 15 nevi). 

\section{Quantification of domain shift}
\label{sec:domainshift_quantification}

Effective performance of UDA methods requires estimating and reducing the divergence of the source and the target domain \citep{ben2010theory}. That is because performance drop of a classification model on an external test set is related to the divergence between domains \citep{elsahar2019annotate,stacke2020measuring}. Possible domain shifts in dermoscopic datasets are discussed in \autoref{sec:sampling}. As can be seen in \autoref{fig:derma}, melanoma often appear darker overall and may look different in general. The imbalanced class distribution and appearence of melanoma being darker overall may affect the results of the quantification measures. To avoid any risks of affection, we quantified domain shifts for the two classes, melanoma and nevus, separately. As it is essential how machine learning models perceive these domain shifts, we analyzed the outcome of two divergence metrics: Cosine similarity and JS divergence. We also visualized the t-SNE projections of the datasets to evaluate the feature space distribution of the data.

\subsection{Divergence metrics}
\label{ssec:div_metrics}

\begin{figure}[h!]
\begin{minipage}[b]{1.0\linewidth}
  \centering
  \centerline{\includegraphics[width=10.5cm]{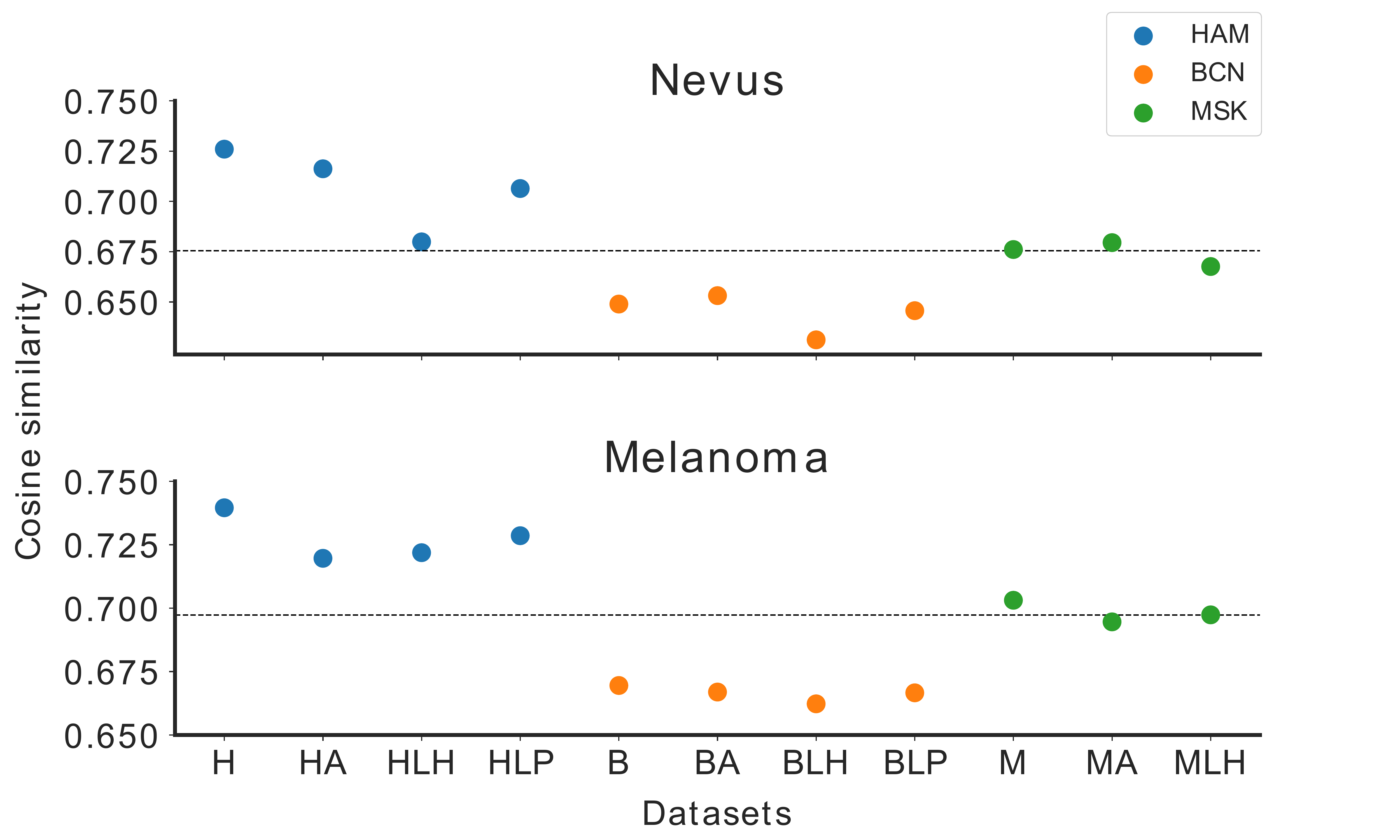}}
  \medskip
\end{minipage}
\caption{Cosine Similarity for 30 iterations per dataset for nevus and melanoma. A high value represents higher similarity. It shows overall low standard deviation. For dataset H we do not expect a domain shift. Dashed line is the mean of all runs. Abbreviations of the datasets are given in \autoref{table:samplingtable1}.}
\label{fig:cossim}
%\end{figure}

%\begin{figure}[h!]
\begin{minipage}[b]{1.0\linewidth}
  \centering
  \centerline{\includegraphics[width=10.5cm]{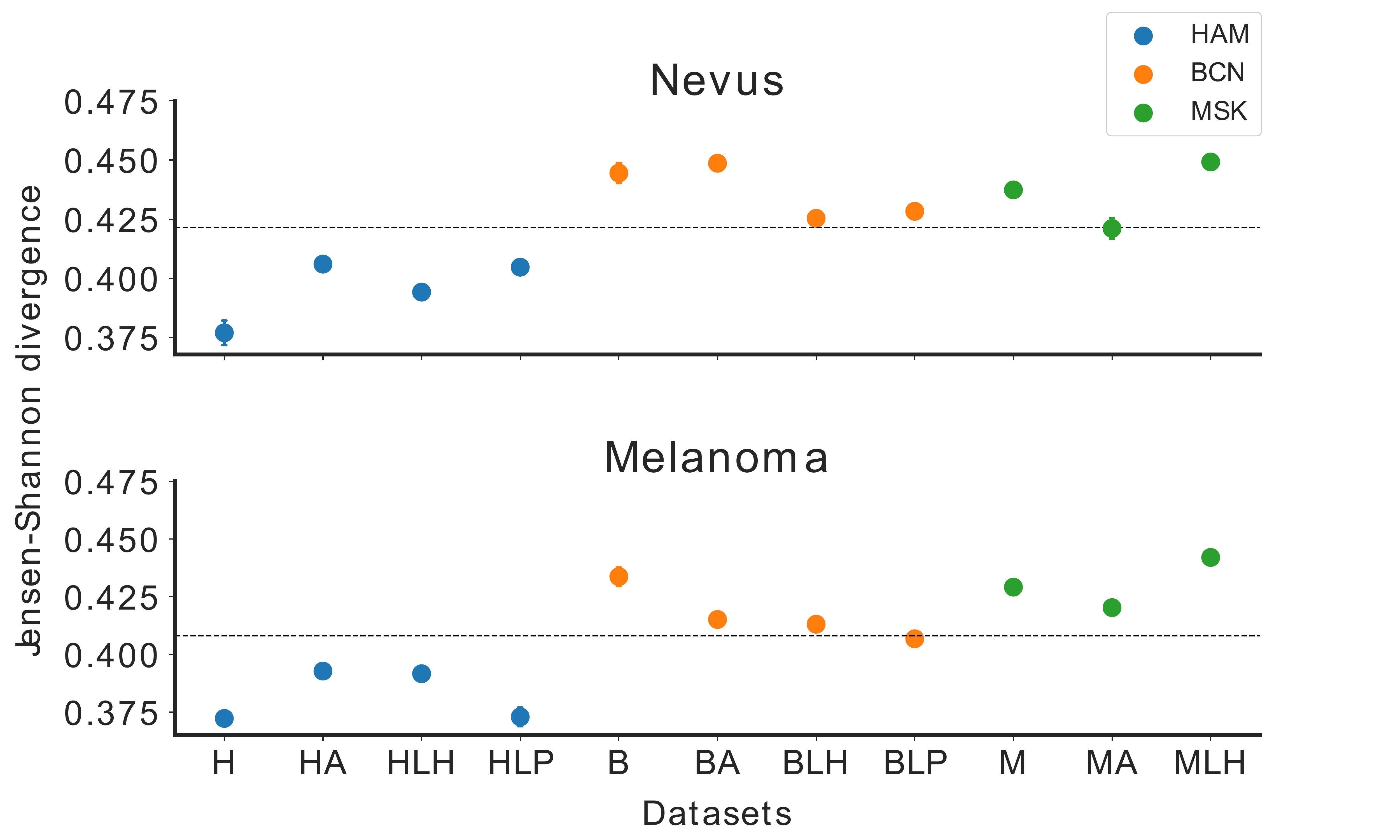}}
  \medskip
\end{minipage}
\caption{JS divergence for 30 iterations per dataset for nevus and melanoma. A low value represents higher similarity. It shows overall low standard deviation. For dataset H we do not expect a domain shift. Dashed line is the mean of all runs. Abbreviations of the datasets are given in \autoref{table:samplingtable1}.}
\label{fig:JSdivergence}
\end{figure}

As discussed in \autoref{sec:Relevant}, divergence metrics can be estimated in both, image- and feature-space. Primarily, we tried to evaluate the similarity of images in the feature space by using \textbf{Cosine similarity}. However, as the image features are extracted from a model, e.g. from the last hidden layer of a ResNet18 model pretrained on ImageNet, estimating this similarity of images in feature space is model-dependent. To overcome this limitation, we additionally calculated \textbf{Jensen Shannon (JS) divergence} in image space. These two measures have already been applied in the medical domain \citep{palladino2020unsupervised, allegretti2021supporting} 

Both metrics are estimated by comparing H dataset with each of the other grouped domain shifted datasets in \autoref{table:samplingtable1}. The specifics for the quantification of Cosine similarity and JS divergence are the same, except that the first is calculated in feature space, while the second is measured in image space. Both metrics are computed pairwise for a random sample of 250 images per dataset. This experiment is repeated 30 times with replacement, while calculating mean, median and standard deviation of the metrics. In order to select an optimal number of image samples for these metrics we conducted experiments with a varying number of samples (\autoref{fig:cossim_samples}, \autoref{fig:JSdivergence_samples}). Although the ideal way of measuring divergence between datasets is to compare each combination of the samples, this approach is computationally intensive. Apart from this, we are mainly interested in a relative separation between the datasets to identify domains in dermoscopic datasets.

In \autoref{fig:cossim_samples} and \autoref{fig:JSdivergence_samples} it can be noticed that beyond a sample size of 250 images the results become more stable over all 30 runs, showing a smaller standard deviation. We verified this effect on small (MLH dataset of size 292) and on large (B dataset of size 4639) datasets. Consequently we selected a sample size of 250 images to quantify the divergence metrics resulting in \autoref{fig:cossim} and \autoref{fig:JSdivergence}. The diagrams show, that melanoma and nevus have a mainly similar distribution pattern. It can also be noticed in both metrics that there exists a clear separation of HAM datasets to BCN and MSK. Moreover, BCN datasets show larger domain shifts with respect to H dataset than MSK datasets. Overall, the standard deviation of all measures over 30 iterations is very low, which indicates a stable quantification.

\subsection{Domain discriminator as quantification measure}
\label{ssec:dd}

A domain discriminator is the first and usual approach to measure the shift between two domains \citep{elsahar2019annotate}.  However, it cannot be used reliably as a primary quantification metric to evaluate domain shifts in ISIC archive due to the presence of duplicates in the datasets \citep{cassidy2022analysis}. Nevertheless, we implemented a domain discriminator to evaluate if it can still differentiate between the datasets from different domains and also if the results correlate with the divergence metrics discussed in \autoref{sec:domainshift_quantification}. In this process, we train a ResNet50-based classifier pretrained on ImageNet to differentiate between a source (always H) and a target dataset. If the resulting performance is low, the discriminator is not able to differentiate well, which indicates a small domain shift. On the contrary, a high performance indicates a large domain shift. For this purpose, source and target datasets are combined and split into a new training and test set (testsize=0.25). 

As can be seen in \autoref{table:samplingtable1}, the amount of images per class for H (source) is different to the amount of images per class in each of the target datasets, where we have very different dataset sizes overall. To overcome this, we used sampling with replacement (n=100 images) to select an equal amount of samples per dataset. Apart from that, to handle the imbalance of class distributions we included a weighted random sampler to balance the data per batch over 20 epochs. As a baseline, we also tested the domain discriminator on classifying images from the exact same dataset (H dataset).

The performance of the domain discriminator on all domain shifted datasets is represented in \autoref{fig:domain_discriminator}. As expected, the performance between training (H) and holdout (H) dataset is poor, indicating the domain discriminator is not able to distinguish well between images of the same dataset. MSK- and BCN-based datasets are showing a larger domain shift quantified by the domain discriminator.

\subsection{t-SNE visualization}
\begin{figure}[h!]
\centering
\begin{minipage}[b]{0.4\linewidth}
\centerline{\includegraphics[width=7.5cm]{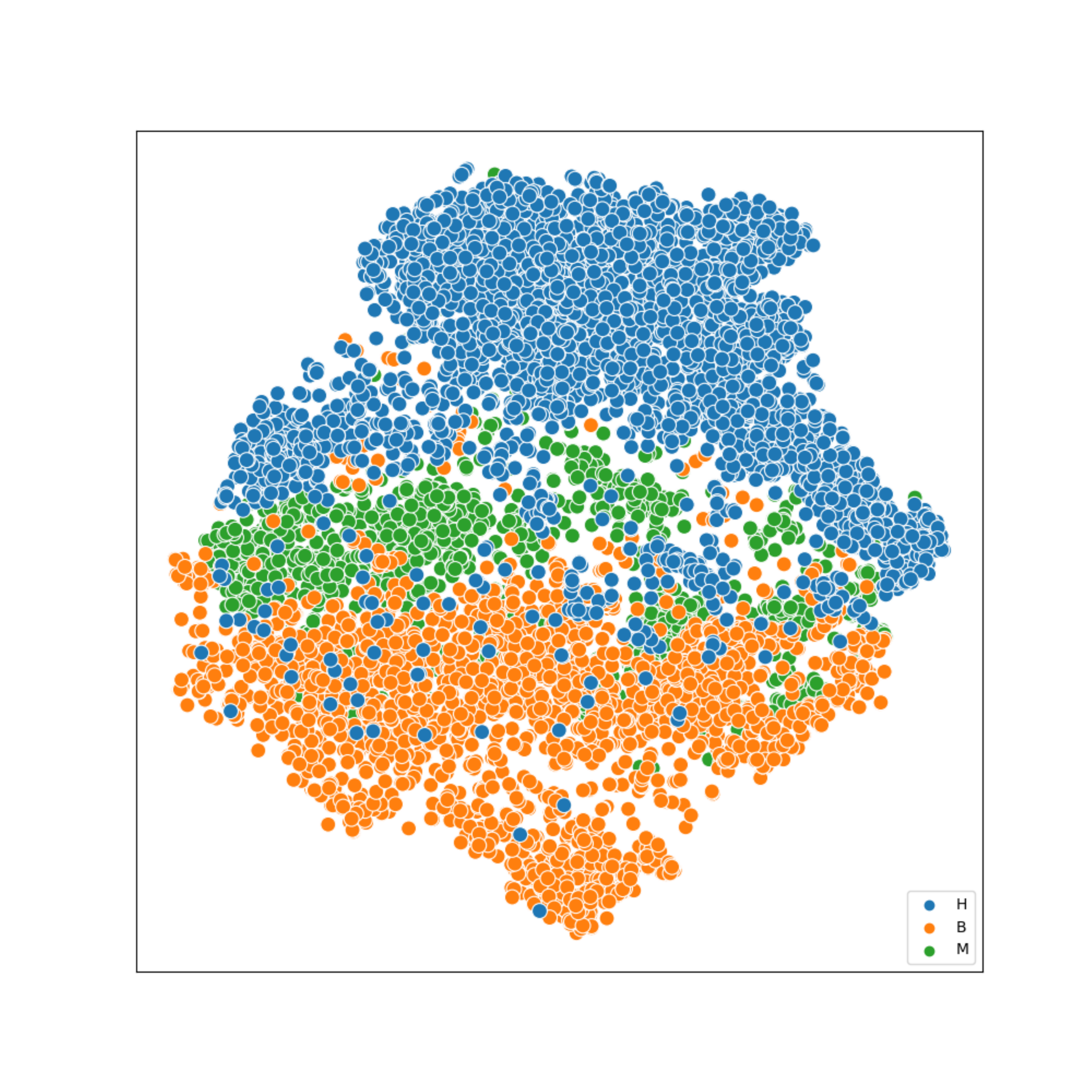}}
\end{minipage}
\hspace{2em}
\begin{minipage}[b]{0.4\linewidth}
\centerline{\includegraphics[width=7.5cm]{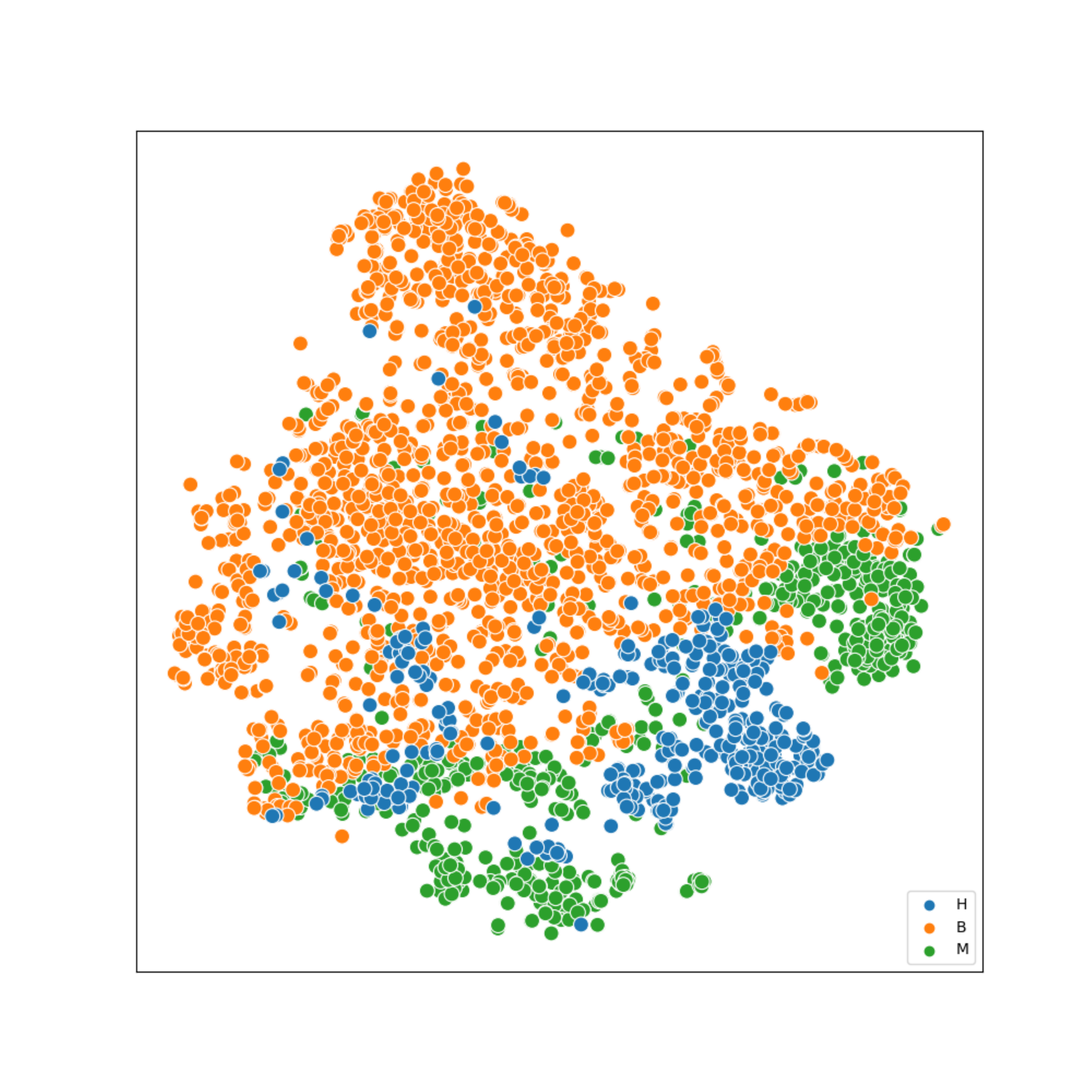}}
\end{minipage}
\caption{t-SNE projection for H, B, M datasets, which represents the technical shifts between the datasets. Abbreviation of the datasets are given in \autoref{table:samplingtable1}.}
\label{fig:tsneplot_def}
\end{figure}

\begin{figure}[h!]
\centering
\begin{minipage}[b]{0.4\linewidth}
\centerline{\includegraphics[width=7.5cm]{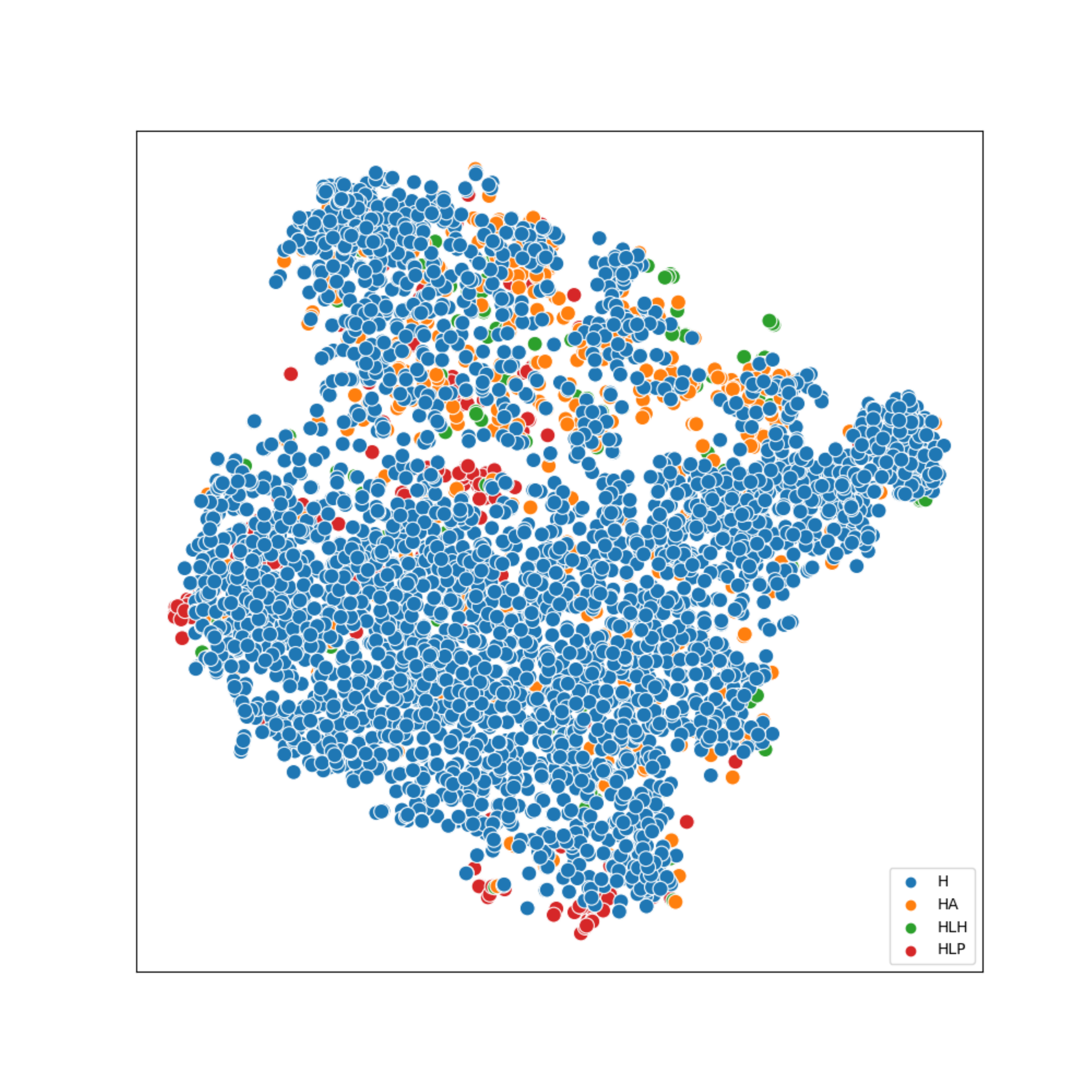}}
\end{minipage}
\hspace{2em}
\begin{minipage}[b]{0.4\linewidth}
\centerline{\includegraphics[width=7.5cm]{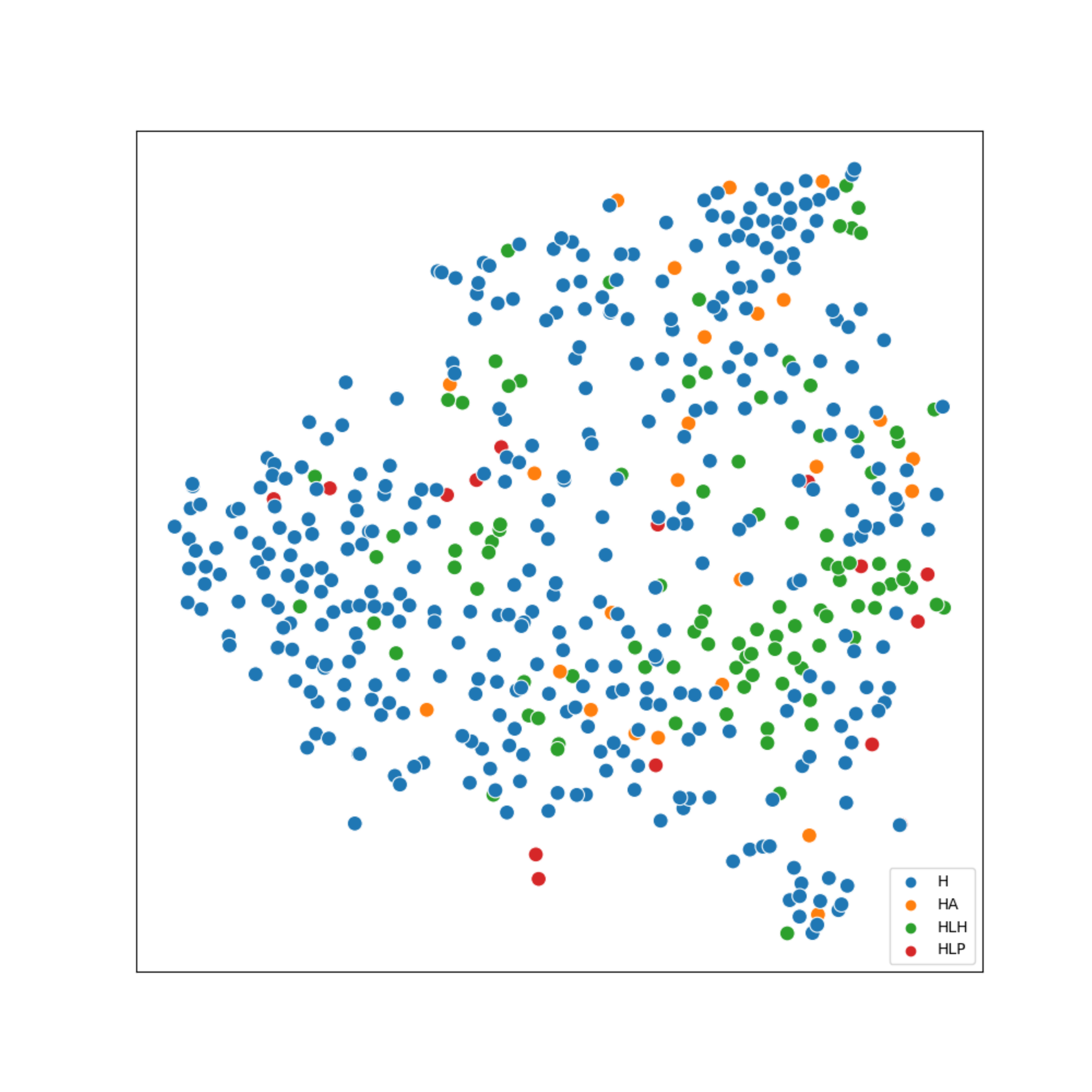}}
\end{minipage}
\caption{t-SNE projection for all HAM datasets representing the biological shifts between datasets. Abbreviation of the datasets are given in \autoref{table:samplingtable1}.}
\label{fig:tsneplot_HAM}
\end{figure}

Apart from the quantification of domain shifts, we also verified how the feature space of different domains is separated in terms of t-SNE projections. For these visualizations, we used all the images in each dataset. The analysis is done for melanoma and nevus separately. \autoref{fig:tsneplot_def} shows the separation of H, B and M datasets (\autoref{table:samplingtable1}) which characterize technical shifts. The figure shows that nevus is more clearly separated compared to melanoma. We have also visualized the domain separation for H with respect to all other datasets originating from HAM (HA, HLH, HLP) (\autoref{fig:tsneplot_HAM}). Biological shifts are not as clearly separated as the technical shifts. These results agree with the divergence metrics shown in \autoref{fig:cossim} and \autoref{fig:JSdivergence}. Furthermore, we visualized t-SNE projections for BCN datasets (B, BA, BLH, BLP) (\autoref{fig:tsneplot_BCN}) and MSK datasets (M, MA, MLH) (\autoref{fig:tsneplot_MSK}) with respect to H dataset. 

Overall, the results of divergence measures agree with the t-SNE visualizations. We further use these verified domain shifted datasets for testing the performance with and without a UDA method discussed in \autoref{sec:influence}.

\section{Influence of domain shifts on a UDA-method}
\label{sec:influence}

To analyze what influence different kinds of domain shifts have, we compared the performances of a ResNet50 with and without an unsupervised domain adaptation method while using our grouped domain shifted datasets. By comparing source only with one adaptation method we can show in what cases and for what kind of datasets or shifts the performance improves and for what cases it does not. 

Various unsupervised domain adaptation methods have been developed, focusing on e.g. discrepancy-based, adversarial-based and reconstruction-based methods \citep{madadi2020deep}. Some of these methods have been tested on different domain adaptation datasets, although we did not find any relevant literature for dermoscopic datasets. The UDA method used in this experiment is based on a \textit{domain-adversarial neural network} (DANN), which showed good results for the adaptation of digit image datasets \citep{ganin2015unsupervised, ganin2016domain}. Dermoscopic skin cancer classification is a much more complex case, where it is even difficult for dermatologists to differentiate between benign and malignant. Although we have labelled target data available in this scenario, we wanted to focus specifically on an unsupervised adaptation technique as it may be more useful for real-world use cases.

\begin{figure}[h!]
\begin{minipage}[h!]{1.0\linewidth}
  \centering
  \centerline{\includegraphics[width=11.5cm]{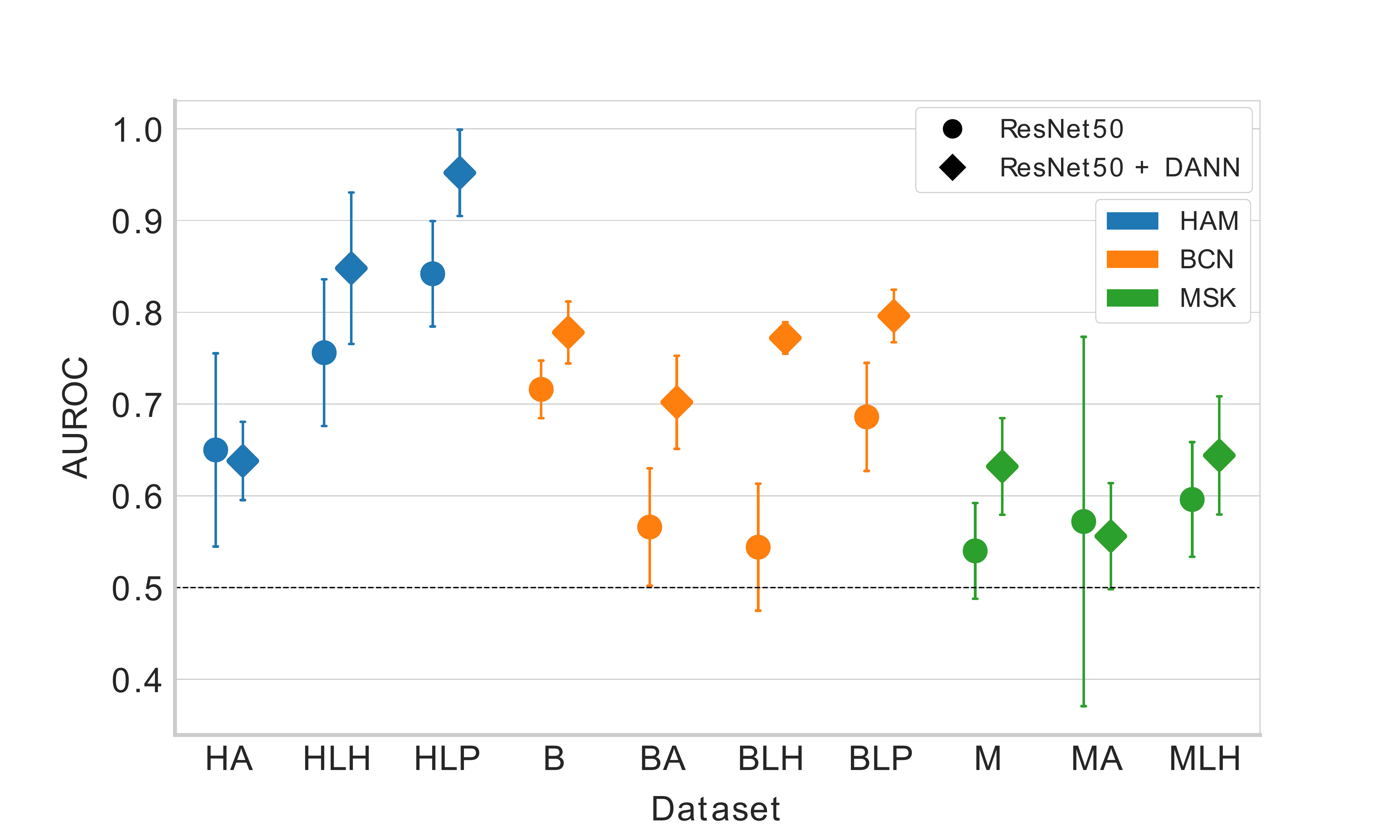}}
\end{minipage}
\caption{Classification of melanoma and nevus. Comparison of AUROC-performance for ResNet50 with and without DANN for different domain-shifted datasets. Abbreviations of the datasets are given in \autoref{table:samplingtable1}.}
\label{fig:dann}
\end{figure}

The network architecture of DANN \citep{ganin2015unsupervised, ganin2016domain} consists of three parts: a feature extractor, a label classifier and a domain classifier. The first two represent the regular feed forward procedure. The domain classifier is added to discriminate between both domains (source and target). Weights are updated by a negative lambda-parameter during backpropagation, so the domain classifiers loss is increasing, while the label classifiers loss is decreasing as usual. Discriminativeness is established with the label classifier and domain-invariance is ensured by the domain classifier.

For the experiments we used a ResNet50 for feature extraction, which is a good performing backbone for CV tasks in general \citep{he2016deep}. However, the choice of a backbone is not relevant in this case, as it is only used for feature extraction for both source-only (ResNet50) and the UDA method. Therefore it is exchangeable. The training-holdout split was performed by stratifying the label classes (melanoma, nevus) in H dataset (\autoref{table:samplingtable1}) to ensure an identical label distribution in both sets. The training consisted of 80\%, while the holdout set consisted of 20\% of the data. We conducted our training process using a pre-existing repository\footnote{\texttt{https://github.com/thuml/Transfer-Learning-Library}}, without any modifications to its hyperparameters. We adapted the library to suit our dermoscopic datasets by making changes to the data loading process. Our experiments were conducted using five different seeds.

As can be seen in \autoref{fig:dann}, DANN improves the performance for almost every dataset (8 out of 10), except for HA (HAM age $\leq30$) and MA (MSK age $\leq30$). We expected the results of the source-only model to be worse than with a DANN, as the major limitation of existing models is to adapt to new and unseen domains. The results show, that DANN can indeed generally improve the performance on domain shifted datasets.

\section{Discussion}
\label{sec:discussion}

As discussed in \autoref{ssec:sampling_procedure}, we used H dataset as the training data for comparing and analyzing the domain shifts present in HAM, BCN and MSK datasets. We chose H (\autoref{table:samplingtable1}), because it represents the typical patient group with skin lesions in clinics, where the patients are usually older than 30 years and the localization of the lesion is either on the torso or on the extremities. All the other grouped datasets are used as real world domain shifts, because they represent less frequent and more specific skin lesion cases, which are not seen in the clinic every day, e.g. younger patients. But also for these rare cases we want classifiers used in the clinic to be able to classify with the same accuracy as typical cases. 

\subsection{Characterization of domain shifts}
\label{ssec:group_discussion}

Presumably, there are many more new conditions than acquisition system settings, age and localization, which can be considered a domain shift. Our datasets provide a starting point in this direction. In further works, specifically artefacts in dermoscopic images, as rulers and black borders in the images and hair on the skin, could be considered for domain shift analysis and domain adaptation tasks. Although we excluded gender as a different domain (as explained in \autoref{sec:sampling}), it could be worthwhile to investigate. Male and female skin has different thickness and hair growth, so the background of these images could look slightly different. Also male have more melanoma on torso, while female usually get melanoma on the lower extremities. As there will probably be more male patients than female with torso melanoma, especially this case could be seen as a new rare domain.

For the translation into the clinic, it would also be relevant to find and generate domain shifted datasets for multi-class classification problems with a sufficient amount of data. Especially rare subgroups of skin lesions can probably be seen as domain shifted datasets. Also, the grouped datasets do not need to be limited to classification tasks only. The grouping procedure could also be applied to other public or private dermoscopic datasets. Apart from this, the grouping method in \autoref{fig:sampling_procedure} could be further extended, e.g. for lesion diameters or ABCDE rules \citep{duarte2021clinical} to detect risky lesions. 

\subsection{Utilization of quantification measures}
\label{ssec:quantification_discussion}

To prove that our grouped datasets are indeed domain shifted, we additionally focused on quantification measures for verification. In cases, where the domain shifts can obviously be recognized with the human-eye, as for instance, in the digit example (e.g. MNIST), it is easy to simply anticipate that they are shifted, if the digits are rotated or coloured. But in dermoscopic datasets, where even experienced doctors find it difficult to differentiate between melanomas and nevi, we cannot simply anticipate a domain shift. Hence, we evaluated multiple  quantification metrics to measure the domain shift between datasets. We provide multiple examples, of how domain shifts can be measured, confirm the shift and also its intensity in the datasets. Our quantification results show, that all BCN and MSK originated datasets are considered domain shifted (w.r.t H dataset) by all measures. We can also observe that biological shift exists and has an effect. If there would only be technical shifts present in the datasets, the default sets (B, M) would show an equal domain shift than BA, BLH, BLP, MA and MLH. So we can confirm, that datasets which include a technical and a biological shift have a higher domain shift to the source dataset. 

\subsection{Performance improvement with UDA method}
\label{ssec:uda_discussion}

Moreover, we also verified the applicability of one UDA method on dermoscopic datasets, which is widely used for other domain adaptation tasks. Even though a domain adaptation method works well for one dataset, it might not result in a similar performance for another. However, for domain shifted dermoscopic datasets the UDA method appears to reduce the domain gap between the datasets while adapting to new domains (\autoref{fig:dann}). We chose DANN because of its popularity and the possibility to be added to any architecture that can be trained with backpropagation \citep{ganin2015unsupervised}. DANN improves the performance for 8 out of 10 dermoscopic datasets. There could be multiple reasons, both biological and technical, why DANN is not able to improve the performance for HA and MA. In these two datasets of young patients also children are included. We did not consider children younger than 15 as a separate domain, as there were only few examples within HAM, BCN and MSK datasets. Children are usually a specific case in clinical diagnosis, where the typical ABCDE features cannot be applied, because the melanomas look differently \citep{haliasos2013dermoscopy, scope2016study}. Both datasets consist of around 15\% underaged patients. Even though BA (BCN age $\leq30$) dataset also consists of around 15\% children. However, for BA dataset DANN seems to improve the performance. Also the distribution of nevi and melanomas in these underaged patients does not show any differences in comparison to the BCN dataset, except that the dataset itself is larger. Another reason for the bad performance could be the presence of inherent bias in the form of artefacts in the images, e.g. black areas. Our domain shift quantifications do not show any significant differences regarding the intensity of the shifts when compared to other datasets.  

\subsection{Correlations between domain shift quantifications}
\label{ssec:correlation}

To observe a potential linear relationship between the previously estimated domain shift quantifications, we measured the Pearson correlation coefficients between the performance drop of a domain discriminator and the divergence metrics, as it was done in other works \citep{elsahar2019annotate, stacke2020measuring}. 

%interpretation
In \autoref{table:corr_mel} and \autoref{table:corr_nev}, we can observe a strong linear relationship between the performance drop of the domain discriminator (AUROC) and the divergence metrics (JS divergence and Cosine similarity) for dermoscopic datasets. In addition to AUROC drop, we also measured the balanced accuracy drop. As both are highly correlated we show only one AUROC for the comparison.  

Cosine similarity negatively and JS divergence positively correlates with the performance drop of the domain discriminator. Therefore Cosine similarity and JS divergence have a strong negative correlation - what we already assumed in \autoref{sec:domainshift_quantification}.

\section{Conclusions}

As, to our knowledge, no benchmark data for domain adaptation on dermoscopic images exists, we suggest a procedure to group images from the publicly available ISIC archive. These grouped datasets can be used to test the generalization capabilities of different models on domain shifted dermoscopic datasets. Along with the domain shifted datasets, we also provide our data grouping procedure, which serves as a starting point for the generation of further dermoscopic datasets for domain adaptation. Additionally, we quantified possible domain shifts in between the different grouped datasets to verify if a shift in fact exists. Our measured divergence metrics agreed with the t-SNE-based visualizations. Finally, we analyzed and compared the performance of a ResNet50-classifier with and without an unsupervised domain adaptation method on all grouped datasets. DANN improves the performance compared to a source-only model in 8 out of 10 datasets and can therefore be used for better generalizability.

\newpage
\section*{Acknowledgements}
This research is funded by the \textit{Helmholtz Artificial Intelligence Cooperation Unit} [grant number ZT-I-PF-5-066].

The Helmholtz AI funding enabled the close cooperation between DKFZ and DLR, which leads to an interdisciplinary exchange between two research groups and thereby enables the integration of novel perspectives and experiences.

%%Harvard
%\bibliographystyle{model2-names.bst}\biboptions{authoryear}
\bibliographystyle{abbrvnat}
\bibliography{main}
\onecolumn
\listoffigures
%%%%%%%%%%%%%%%%%%%%%%%%%%%%%%%%%%%%%%%%%%%%%%%%%%%%%%%%%%%%%%%%%%%%%%%%%%%%%%%%%%%%%%%%%%%%%%%%%%%%%

%\section*{References}
%
%Please ensure that every reference cited in the text is also present in
%the reference list (and vice versa).

%\section*{\itshape Reference style}
%
%Text: All citations in the text should refer to:
%\begin{enumerate}
%\item Single author: the author's name (without initials, unless there
%is ambiguity) and the year of publication;
%\item Two authors: both authors' names and the year of publication;
%\item Three or more authors: first author's name followed by `et al.'44
%and the year of publication.
%\end{enumerate}
%references should be listed first alphabetically, then chronologically.%Citations may be made directly (or parenthetically). Groups of
\newpage

%\appendix

%\onecolumn
\section*{Appendix}
%\vspace{-15pt}
\setcounter{figure}{0} 
\label{ssec:isic2}
\begin{figure*}[h!]
\begin{minipage}[b]{1.0\linewidth}
  \centering
  \centerline{\includegraphics[scale=0.5]{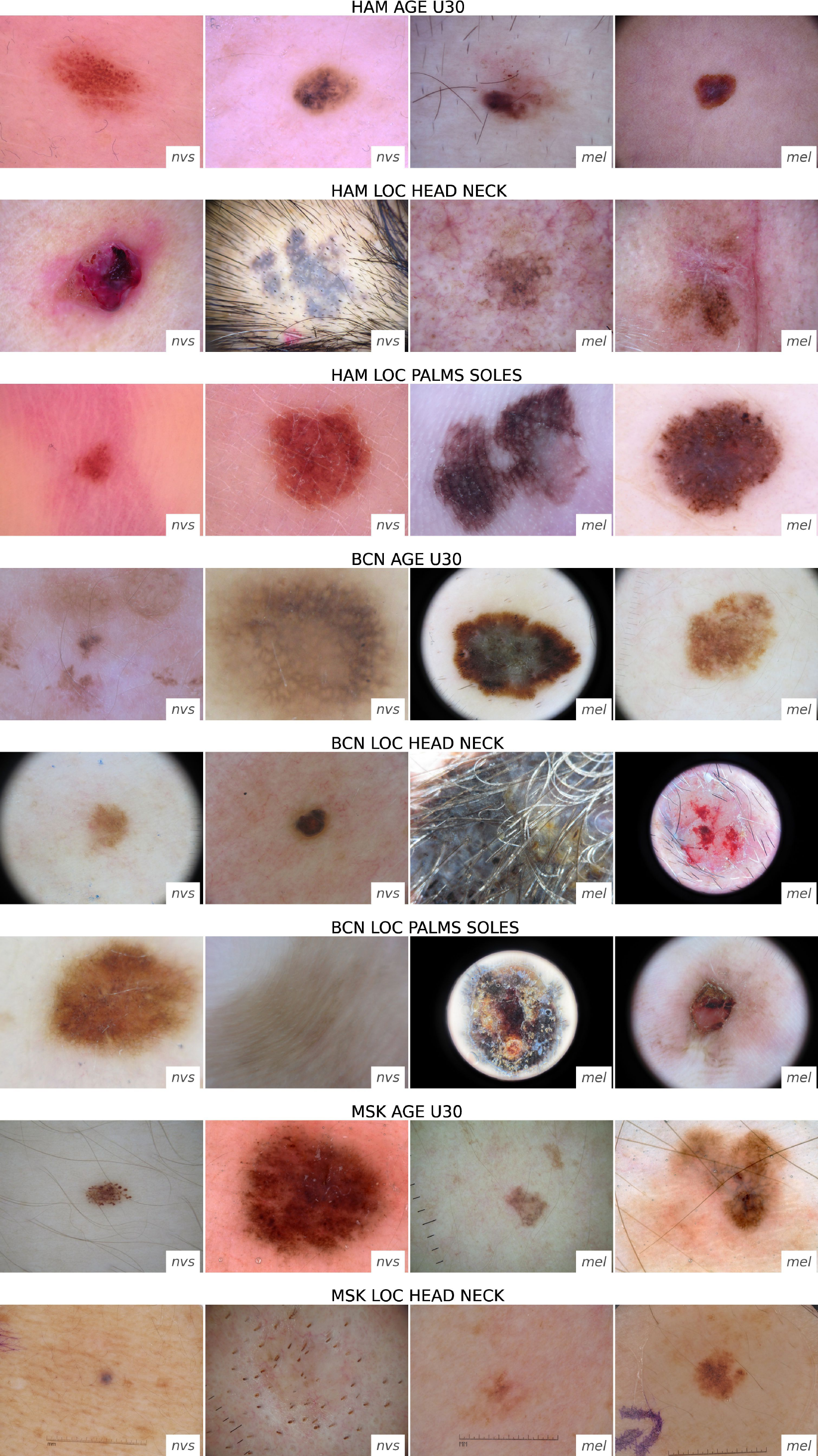}}
  \caption{Randomly selected dermoscopic images stratified by label per grouped dataset.}
\label{fig:derma2}
\end{minipage}
\end{figure*}

\begin{figure*}[h!]
\begin{minipage}[b]{1.0\linewidth}
  \centering
  \centerline{\includegraphics[width=14cm]{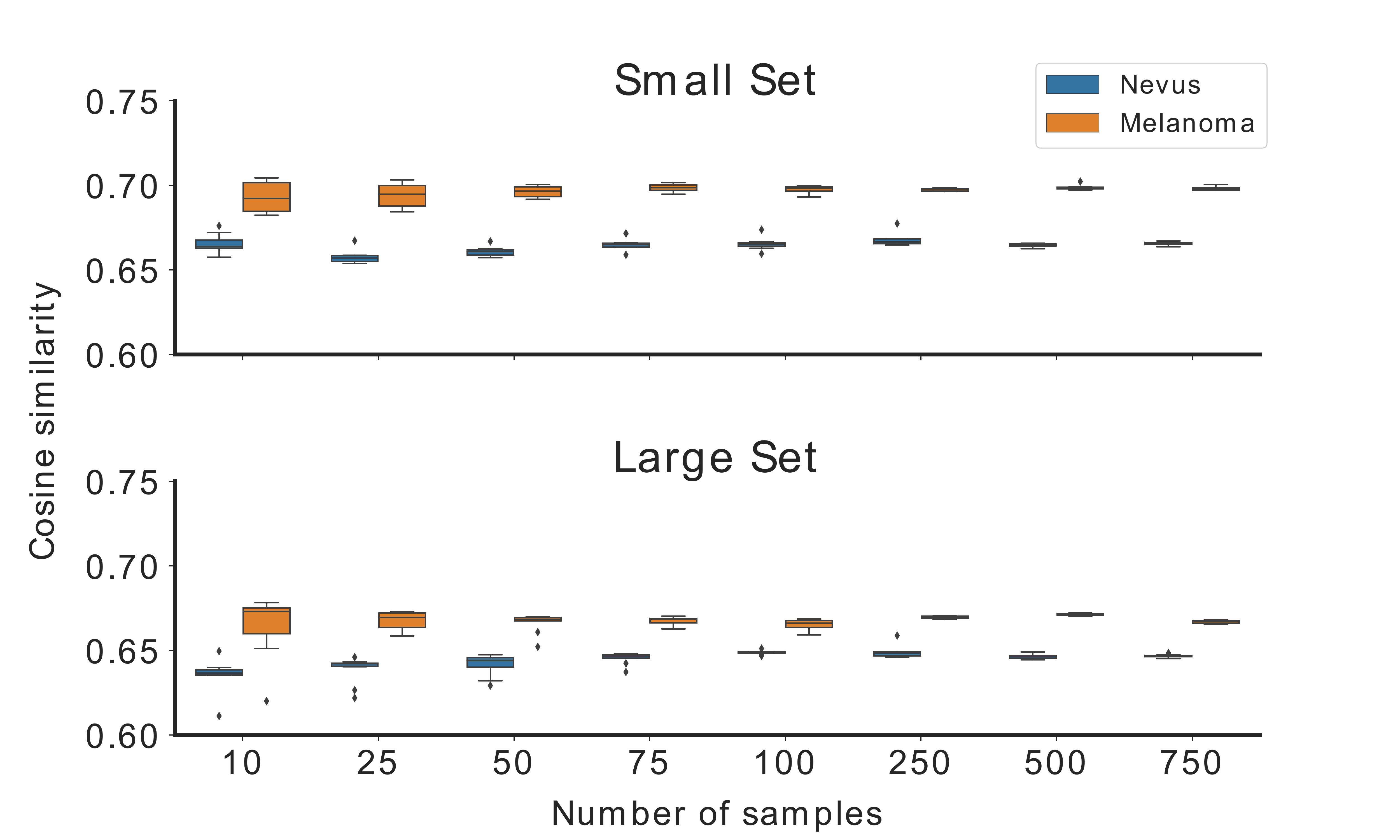}}
\end{minipage}
\caption{Cosine Similarity for a different number of samples. Small dataset is MLH (n=292), large dataset is B (n=4639).}
\label{fig:cossim_samples}
\end{figure*}

\begin{figure*}[h!]
\begin{minipage}[b]{1.0\linewidth}
  \centering
  \centerline{\includegraphics[width=14cm]{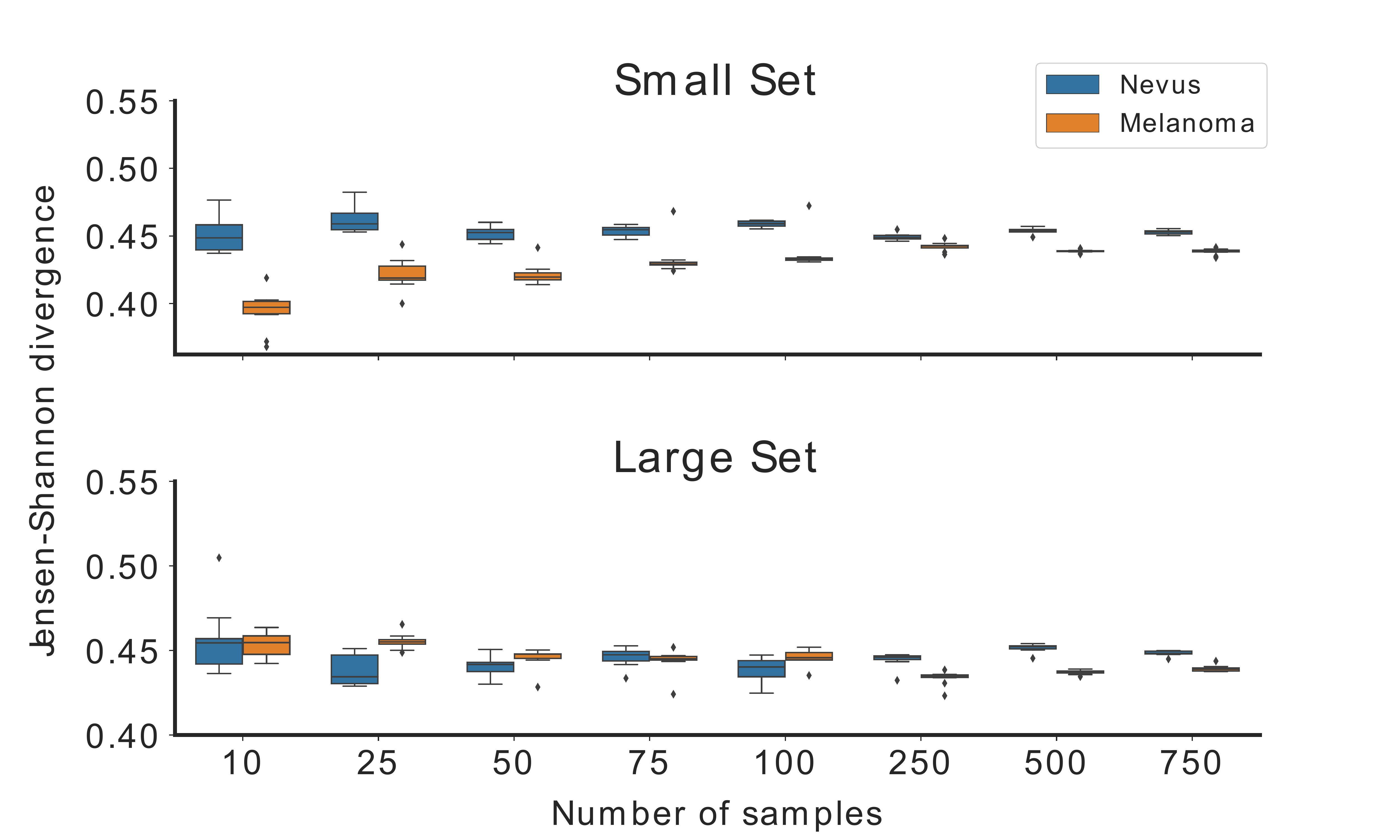}}
\end{minipage}
\caption{JS divergence for a different number of samples. Small dataset is MLH (n=292), large dataset is B (n=4639).}
\label{fig:JSdivergence_samples}
\end{figure*}

%\twocolumn

\begin{figure}[h!]
\centering
\begin{minipage}[b]{0.4\linewidth}
\centerline{\includegraphics[width=7.5cm]{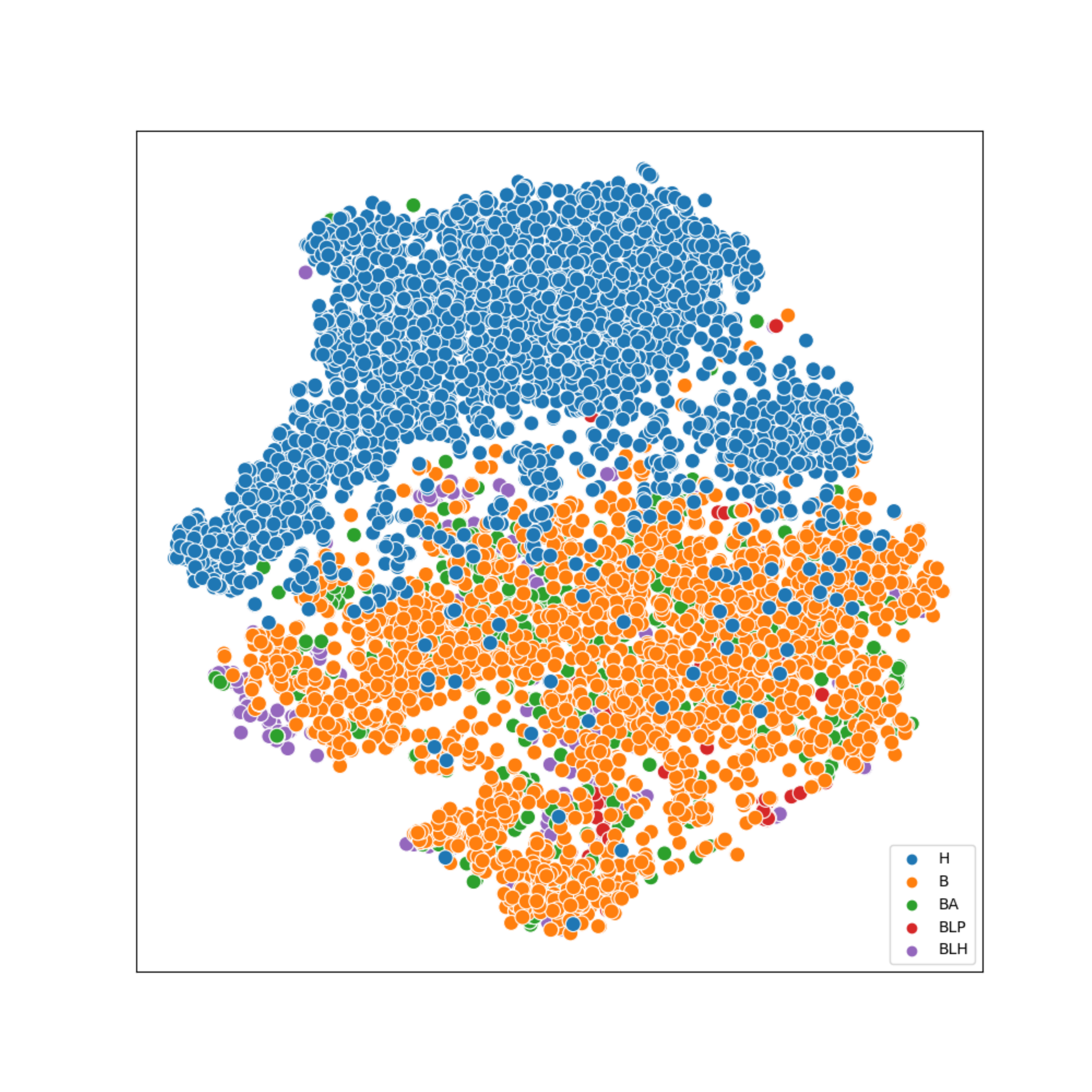}}
\end{minipage}
\hspace{2em}
\begin{minipage}[b]{0.4\linewidth}
\centerline{\includegraphics[width=7.5cm]{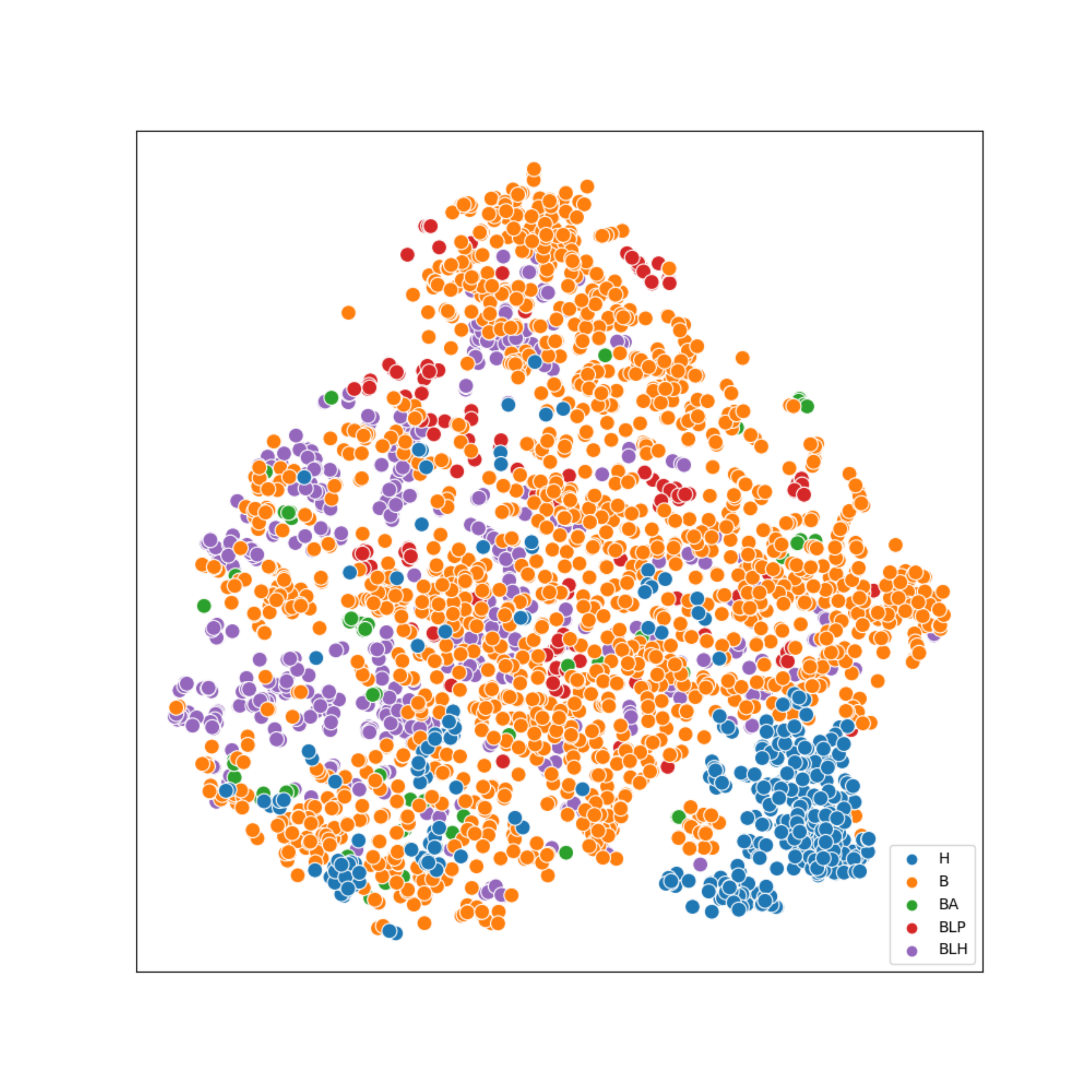}}
\end{minipage}
\caption{t-SNE projections for H and BCN datasets, which show the technical and biological shifts between the datasets. The top plot shows the separation between datasets in the melanoma class. The bottom plot represents the nevus class. Abbreviations of the datasets are given in \autoref{table:samplingtable1}.}
\label{fig:tsneplot_BCN}
\end{figure}

%\pagebreak
\begin{figure}[h!]
\centering
\begin{minipage}[b]{0.4\linewidth}
\centerline{\includegraphics[width=7.5cm]{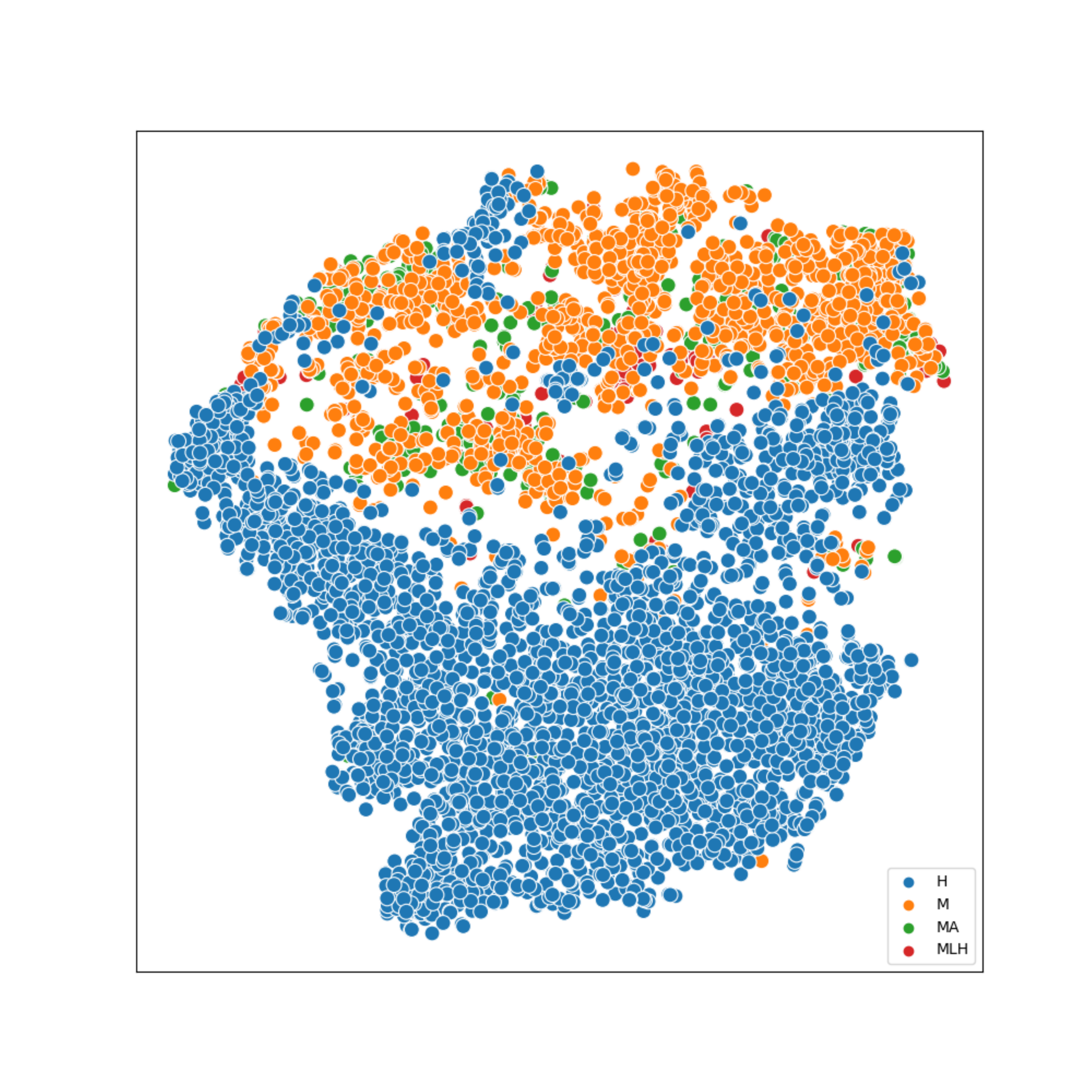}}
\end{minipage}
\hspace{2em}
\begin{minipage}[b]{0.4\linewidth}
\centerline{\includegraphics[width=7.5cm]{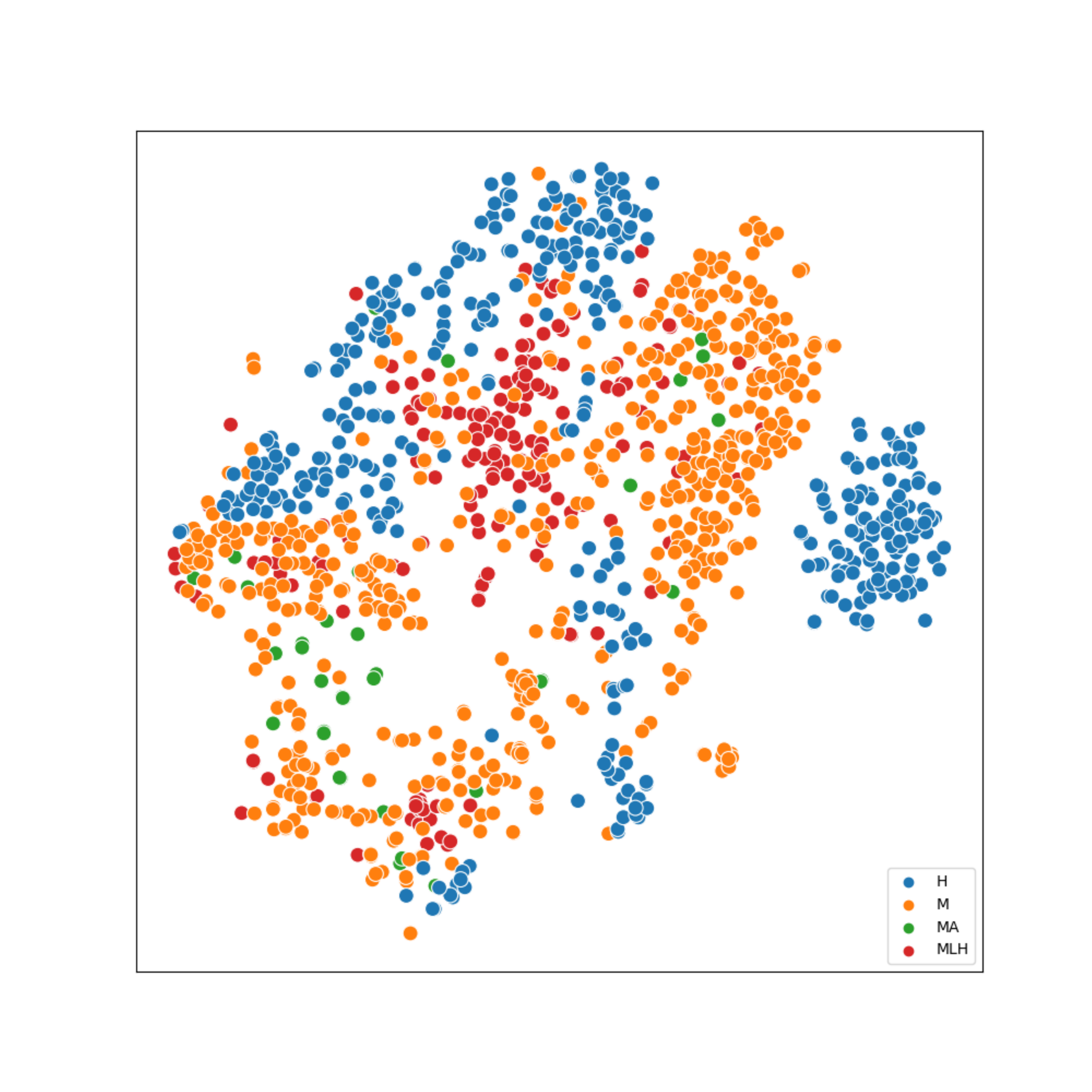}}
\end{minipage}
\caption{t-SNE projections for H and MSK datasets, which represents the technical and biological shifts between the datasets. The top plot shows the separation between datasets in the melanoma class. The bottom plot represents the nevus class. Abbreviations of datasets are given in \autoref{table:samplingtable1}.}
\label{fig:tsneplot_MSK}
\end{figure}

\begin{figure}[ht!]
\begin{minipage}[ht!]{1.0\linewidth}
  \centering
  \centerline{\includegraphics[width=10.5cm]{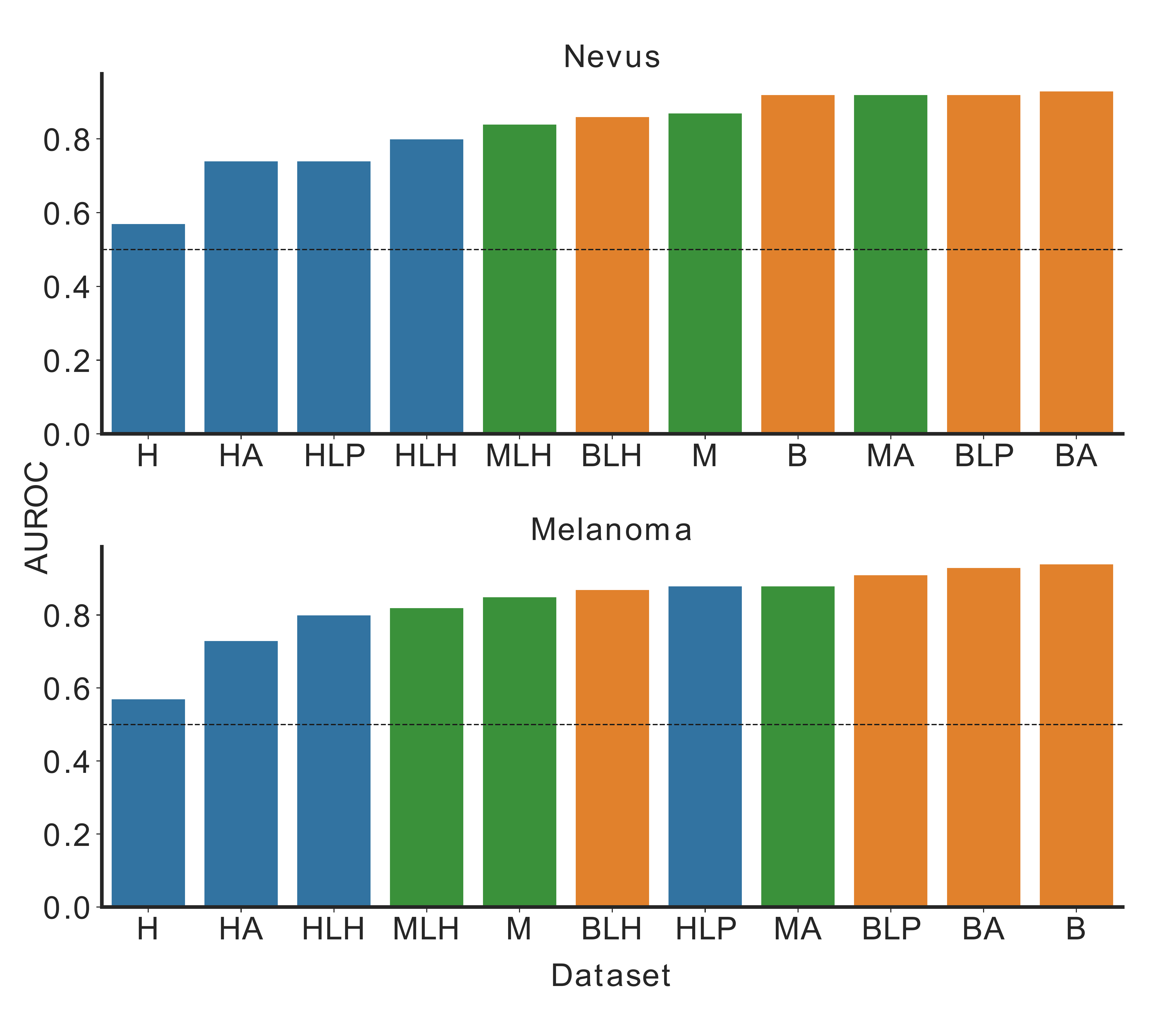}}
\end{minipage}
\caption{AUROC scores of the domain discriminator for each domain shifted dataset with respect to HAM default dataset. Abbreviation of the datasets are given in \autoref{table:samplingtable1}.}
\label{fig:domain_discriminator}
\end{figure}

\setcounter{table}{0} 
\begin{table}[h!]
\centering
\begin{tabular}{c | c c c}
 & JS divergence & Cosine similarity & AUROC drop\\
\hline
JS divergence & 1 & -0.67 & 0.44\\  
Cosine similarity & -0.67 & 1 & -0.71\\ 
AUROC drop & 0.44 & -0.71 & 1 \\
\end{tabular}
\caption{Melanoma - Pearson correlation between performance drop and divergence measures.}
\label{table:corr_mel}
%\end{table}

\vspace{3em}
%\begin{table}[ht!]
\centering
\begin{tabular}{c | c c c}
 & JS divergence & Cosine similarity & AUROC drop\\
\hline
JS divergence & 1 & -0.60 & 0.77\\  
Cosine similarity & -0.60 & 1 & -0.76\\ 
AUROC drop & 0.77 & -0.76 & 1 \\
\end{tabular}
\caption{Nevus - Pearson correlation between performance drop and divergence measures.}
\label{table:corr_nev}
\end{table}

%Supplementary material that may be helpful in the review process should
%be prepared and provided as a separate electronic file. That file can
%then be transformed into PDF format and submitted along with the
%manuscript and graphic files to the appropriate editorial office.    

\end{document}